\documentclass[remotesensing,article,submit,moreauthors,pdftex,10pt,a4paper]{mdpi}

\usepackage{graphicx, subfig}
\usepackage{amssymb}
\usepackage{array}
\usepackage{arydshln}
\usepackage{multirow}
\usepackage{bbm}
\usepackage{textcomp}
\usepackage{xcolor}

\definecolor{citecol}{rgb}{0,0,0.5}
\hypersetup{urlcolor=citecol,linkcolor=black,citecolor=citecol,colorlinks=true}

\newcommand{\ie}{\textit{i.e.} }
\newcommand{\eg}{\textit{e.g.} }
\newcommand{\codeRepo}{\url{https://github.com/charlotte-pel/temporalCNN}}

\definecolor{wheat}{RGB}{255, 255, 0}
\definecolor{barley}{RGB}{255, 255, 220}
\definecolor{rapeseed}{RGB}{241, 138, 217}
\definecolor{corn}{RGB}{255, 165, 0}
\definecolor{soy}{RGB}{255, 51, 51}
\definecolor{sunflower}{RGB}{130, 0, 130}
\definecolor{sorghum}{RGB}{120, 50, 20}
\definecolor{pea}{RGB}{200, 100, 200}
\definecolor{grassland}{RGB}{180, 255, 180}
\definecolor{deciduous}{RGB}{0, 140, 0}
\definecolor{conifer}{RGB}{0, 80, 0}
\definecolor{water}{RGB}{0, 0, 150}
\definecolor{urban}{RGB}{255, 100, 100}

\firstpage{1} 
\articlenumber{x}
\doinum{10.3390/------}
\pubvolume{xx}
\pubyear{2019}
\copyrightyear{2019}
\externaleditor{Academic Editor: name}
\history{Received: date; Accepted: date; Published: date}
\preto{\abstractkeywords}{\nolinenumbers}

\Title{Temporal Convolutional Neural Network for the Classification of Satellite Image Time Series}

\Author{Charlotte~Pelletier $^{1}$*, Geoffrey~I.~Webb $^{1}$ and Fran\c{c}ois~Petitjean $^{1}$}

\address{%
$^{1}$ \quad Faculty of Information Technology, Monash University, Melbourne,
VIC, 3800\\}

\corres{Correspondence: charlotte.pelletier@monash.edu}

\abstract{New remote sensing sensors now acquire high spatial and spectral Satellite Image Time Series (SITS) of the world. These series of images are a key component of classification systems that aim at obtaining up-to-date and accurate land cover maps of the Earth's surfaces. More specifically, the combination of the temporal, spectral and spatial resolutions of new SITS makes possible to monitor vegetation dynamics. Although traditional classification algorithms, such as Random Forest (RF), have been successfully applied for SITS classification, these algorithms do not make the most of the temporal domain. Conversely, some approaches that take into account the temporal dimension have recently been tested, especially Recurrent Neural Networks (RNNs).
This paper proposes an exhaustive study of another deep learning approaches, namely Temporal Convolutional Neural Networks (TempCNNs) where convolutions are applied in the temporal dimension. 
The goal is to quantitatively and qualitatively evaluate the contribution of TempCNNs for SITS classification. This paper proposes a set of experiments performed on one million time series extracted from 46 Formosat-2 images. The experimental results show that TempCNNs are 
more accurate than RF and RNNs, that are the current state of the art for SITS classification. We also highlight some differences with results obtained in computer vision, \eg about pooling layers. Moreover, we provide some general guidelines on the network architecture, common regularization mechanisms, and hyper-parameter values such as batch size. Finally, we assess the visual quality of the land cover maps produced by TempCNNs.}


\begin{document}

\section{Introduction}
    
    The biophysical cover of Earth's surfaces --~land cover~-- has been declared as one of the fifty-four Essential Climate Variables \cite{bojinski_2014}. Accurate knowledge of land cover is indeed a key information for environmental researches, and is essential to monitor the effects of climate change, to manage resources, and to assist in disaster prevention. Accurate and up-to-date land cover maps are critical as both inputs to modeling systems --~\textit{e.g.} flood and fire spread models~-- and decision tools to inform public policy makers \cite{feddema_2005}. 

\begin{figure}
        \centering
        \includegraphics[width=0.75\linewidth]{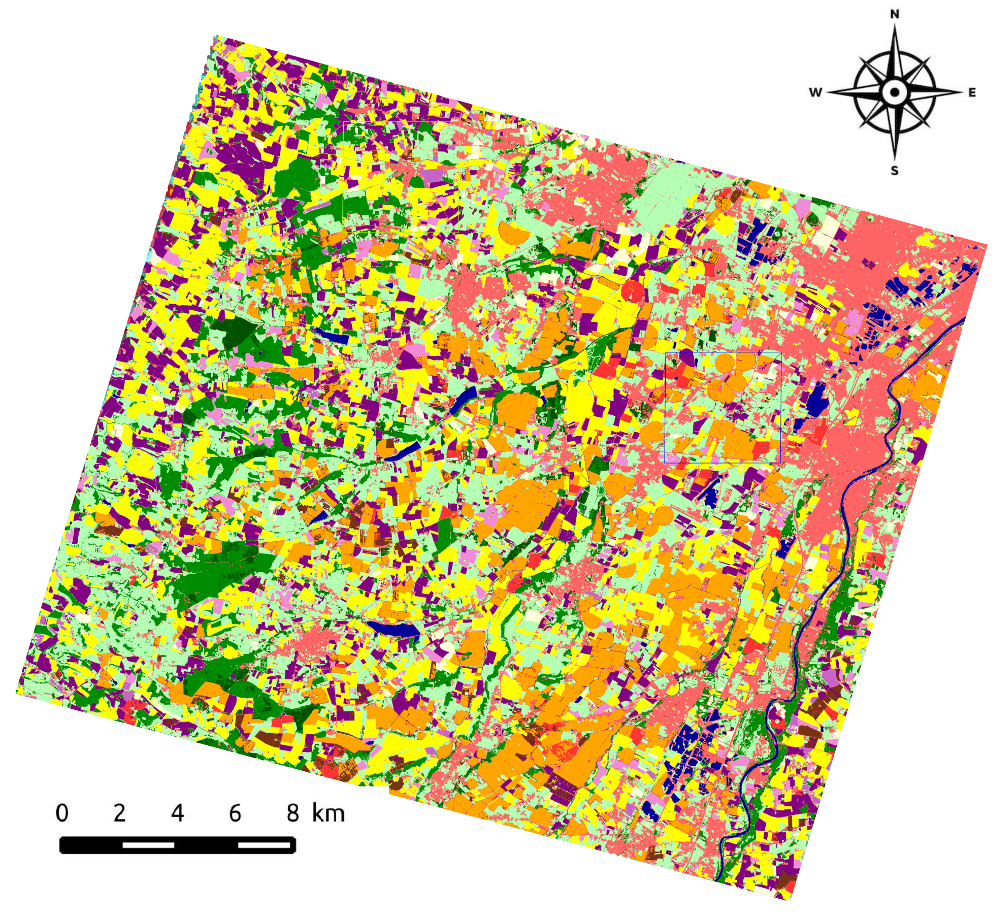}
        \caption{An example land cover map.} 
        \label{fig:lc_map_example}
    \end{figure}

    State-of-the-art approaches to producing accurate land cover maps use supervised classification of satellite images \cite{gomez_2016}. This makes it possible for maps to be reproducible and to be automatically produced at a global scale while reaching high levels of accuracy \cite{inglada_2017}. Figure~\ref{fig:lc_map_example} displays an example of such a map. Latest satellite constellations are now acquiring satellite image time series (SITS) with high spectral, spatial and temporal resolutions. For instance, the two Sentinel-2 satellites provide worldwide images every five days, freely distributed, within 13 spectral bands at spatial resolutions varying from 10 to 60 meters since March 2017 \cite{drusch_2012}. 
    
    These new high-resolution SITS constitute an incredible source of data for land cover mapping, especially for vegetation and crop mapping \cite{matton_2015,vuolo_2018}, at regional and continental scales \cite{immitzer_2016, inglada_2017}. They also question the choice of the classification algorithms: although traditional algorithms showed good performance for SITS classification \cite{gomez_2016}, they do not explicitly use the temporal relationship between acquisitions. Concurrently, deep learning methods are appealing for this task \cite{fawaz_2018}. Their ability to automatically extract useful representations has been demonstrated in machine learning \cite{bengio_2013}, as well as on various remote sensing tasks \cite{zhu_2017}. The recent availability of long SITS on large areas requires now new studies to evaluate if, and how SITS classification task can benefit from deep learning techniques.
    
    In the following, we first review the state-of-the-art approaches used for SITS classification. Then, we describe the most recent applications of deep learning to remote sensing data, including a focus on time series classification. Finally, we present the motivations and contributions of this work.
    
    
    
    
    \subsection{Traditional approaches for SITS classification}
        The state-of-the-art classification algorithms used to produce maps are currently Support Vector Machines (SVMs) and Random Forests (RFs) \cite{khatami_2016}. These algorithms are generally applied at pixel-level on a stack of the multi-spectral images contained in the SITS.
        These algorithms are oblivious to the temporal dimension that structures SITS. The temporal order in which the images are presented has thus no influence on the results: a shuffle of the images in the series would result to the same model, and thus the same accuracy performance. It induces a loss of the temporal behaviour for classes with evolution over time, such as the numerous forms of vegetation that are subject to seasonal change. 

        One solution to mitigate this problem has been to pre-calculate temporal features extracted from vegetation index time series, that are then fed to a classification algorithm \cite{jia_2014, pittman_2010, valero_2016}. Temporal features could be either some statistical values, such as the maximum vegetation index, or the approximation of key dates in the phenological stages of the targeted vegetation classes, \eg the time of peak vegetation index. However, the addition of such temporal features has shown little effect on classification performance \cite{pelletier_2016}.

        To make the most of the temporal domain, other works have applied Nearest Neighbor (NN) -type approaches combined with temporal similarity measures \cite{petitjean_2012}. Such measures aim at capturing the temporal trends present in the series by measuring a similarity independent of some temporal shifts between two time series \cite{maus_2016}. Although promising, these methods require a complete scan of the training set for classifying each test instance. This induces a huge computational complexity, that is prohibitive for applications with more than a few thousand profiles \cite{belgiu_2018}.

    \subsection{Deep Learning in Remote Sensing}

        Deep learning approaches have been successfully used for many machine learning tasks including face detection \cite{schroff_2015}, object recognition \cite{redmon_2017}, and machine translation \cite{bahdanau_2014}. Benefiting from both theoretical and technical advances \cite{krizhevsky_2012,ioffe_2015}, they have shown to be extremely good at making the most of unstructured data such as images, audio, or text. The applications to remote sensing of the two main deep learning architectures --~Convolutional Neural Networks (CNNs) and Recurrent Neural Networks (RNNs)~-- are presented in the following.
        
        \subsubsection{Convolutional Neural Networks}
        
        CNNs have been widely applied to various remote sensing tasks including land cover classification of very high spatial resolution images \cite{maggiori_2017, postadjian_2017}, semantic segmentation \cite{volpi_2017}, object detection \cite{audebert_2017}, reconstruction of missing data \cite{zhang_2018}, or pansharpening \cite{masi_2016}. In these works, CNN models make the most of the spatial structure of the data by applying convolutions in both $x$ and $y$ dimensions. The main successful application of CNNs in remote sensing remains the classification of hyperspectral images, where 2D-CNNs across the spatial dimension have also been tested \cite{liang_2016}, as well as 1D-CNNs across the spectral dimension \cite{hu_2015}, and even 3D-CNNs across spectral and spatial dimensions \cite{li_2017, hamida_2018}. 
      
    
        
        Regarding the classification of multi-source and multi-temporal data, 1D-CNN and 2D-CNN have been used without taking advantage of the temporal dimension \cite{kussul_2017,scarpa_2018}: convolutions are applied either in the spectral domain or in the spatial domain, excluding the temporal one. In other words, the order of the images has no influence on the model and results. 
        
        Meanwhile, temporal 1D-CNNs (TempCNNs) where convolutions are applied in the temporal domain have proven to be effective for handling the temporal dimension for time series classification \cite{wang_2017}, and 3D-CNN for both the spatial and temporal dimension in video classification \cite{wu_2015}. Consequently, these TempCNN architectures, that make the most of the temporal structure of SITS starts to be explored in remote sensing with 1D-CNNs where convolutions are applied across the temporal dimension alone \cite{dimauro_2017,zhong_2019}, and also 3D-CNNs where convolutions are applied in both temporal and spatial dimensions \cite{ji_2018}. In particular, these preliminary works highlights the potential of TempCNNs with highest accuracy than traditional algorithms such as RF. Although similar with the work proposed in \cite{zhong_2019}, this paper proposed an extensive study of TempCNNs (see Section~\ref{subsec:contributions}), and deals with multi-spectral images and a broader nomenclature (not specific in summer crop mapping).
        
        \subsubsection{Recurrent Neural Networks}
        
        RNNs is another type of deep learning architecture that are intrinsically designed for sequential data. For this reason, they have been the most studied architecture for SITS classification. They have demonstrated their potential for the classification of optical time series \cite{russwurm_2017,sun_2018} as well as multi-temporal Synthetic Aperture Radar (SAR) \cite{ienco_2017,minh_2018}. Approaches have been tested with two types of units: Long-Short Term Memory (LSTM) and Gated Recurrent Units (GRUs) showing a small accuracy gain by using GRUs \cite{ndikumana_2018,russwurm_2017}. In addition, these works have shown that RNNs outperform traditional classification algorithms such as RFs or SVMs. 
        
        Some recent works dedicated to SITS classification have also combined RNNs with 2D-CNNs (spatial convolutions) either by merging representations learned by the two types of networks \cite{benedetti_2018} or by feeding a CNN model with the representation learned by a RNN model \cite{russwurm_2018}. These types of combinations have also been used for land cover change detection task between multi-spectral images \cite{lyu_2016, mou_2018}.
        
        RNN models are able to explicitly consider the temporal correlation of the data \cite{ndikumana_2018} making them particularly well-suited to drawing a prediction at each time point, such as producing a translation of each word in a sentence \cite{bahdanau_2014}. In remote sensing, a RNN model based on LSTM units has been proposed to output a land cover prediction at several time steps to detect oil palm plantations \cite{jia_2017}.
        However, land cover mapping usually aims at producing one label for the whole series, where the label holds a temporal meaning (\eg ``corn crop''). RNNs might therefore be less suited to this specific classification task. In particular, the number of training steps (\ie the number of back-propagation steps) is a function of the length of the series \cite[Section 10.2.2]{goodfellow_2016}, while it is only a function of the depth of the network for CNNs. The result is a network that is: 1)~harder to train because patterns at the start of the series are many layers away from the classification output, and 2)~longer to train because the error has to be back-propagated through each layer in turn. 
         
     \subsection{Our Contributions}
     \label{subsec:contributions}
     
        In this paper, we extensively study the use of TempCNNs --~where convolutions are applied in the temporal domain~-- for the classification of high-resolution SITS. The main contributions of this paper include:
        \begin{itemize}
            \item demonstrating the potential of TempCNNs against TempCNNs and RNNS,
            \item showing the importance of temporal convolutions,
            \item evaluating the effect of additional hand-crafted spectral features such as vegetation index one,
            \item exploring the architecture of TempCNNs.
        \end{itemize}
        
        This paper does not propose a unique architecture that should be adopted by practitioners for all SITS classification problems. Rather, we present an experimental study of TempCNNs in order to give general guidelines about how they might be used and parameterized. For this purpose, we will first compare the classification performance of TempCNNs to the one of RFs and RNNs. Then, we will discuss different architecture choices including the size of the convolutions, the pooling layers, the width and depth of the network, the regularization mechanisms and the batch size.
        To our knowledge, this work is the first extensive study of TempCNNs to date. 
        Note that this paper focuses only on 1D-CNN models and does not cover the use of the spatial structure of the data. An interested reader could refer to \cite{ji_2018} for a first exploration of that topic.
            
        All the topics are addressed experimentally using 46 high-resolution Formosat-2 images, with training/test sets composed of one million labeled time series. This paper presents the results obtained over 2,000 deep learning models. It corresponds to more than 2,000 hours of training time performed mainly on NVIDIA Tesla V100 Graphical Processing Units (GPUs).



    This paper is organized as follows: Section~\ref{sec:cnn} describes the TempCNN model used in the experiments. Then, Section~\ref{sec:data} is dedicated to the description of the data and the experimental settings. Section~\ref{sec:exp_results} is the core section of the paper that presents our experimental results. Finally, Section~\ref{sec:clc} draws the main conclusions of this work. 

        
        
        
        

\section{Temporal Convolutional Neural Networks}
\label{sec:cnn}
	
    This Section aims at presenting TempCNN models. First, the theory of neural networks and CNNs is briefly reviewed. Then, the principle of the temporal convolution is presented. Finally, we introduce the general form of the TempCNN architecture studied in Section~\ref{sec:exp_results}.  


    \subsection{General Principles}
    \label{subsec:intro_cnn}   
 
\begin{figure}[!ht]
            \centering 
	        \includegraphics[width=0.55\linewidth]{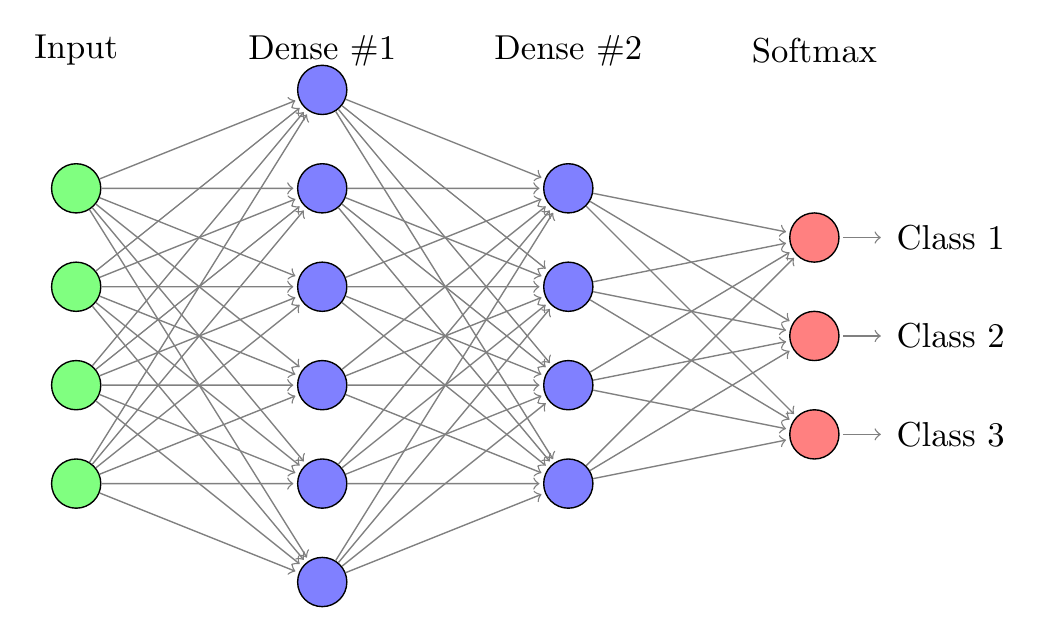}
            \caption{Example of fully-connected neural network.}
            \label{fig:network_examples}
        \end{figure}

        Deep learning networks are based on the concatenation of different layers where each layer takes the outputs of the previous layer as inputs. Figure~\ref{fig:network_examples} shows an example of a  fully-connected network where the neurons in green represent the input, the neurons in blue belong to the hidden layers and the neurons in red are the outputs. As depicted, each layer is composed of a certain number of units, namely the neurons. The input layer size depends on the dimension of the instances, whereas the output layer is composed of $C$ units for a classification task of $C$ classes. The number of hidden layers and their number of units need to be selected by the practitioner.
        
        Formally, the outputs of a layer $l$, the activation map denoted by $A^{[l]}$, are obtained through a two-step calculation: it first takes a linear combination of the inputs --~which are the output of layer $l-1$, \ie{}$A^{[l-1]}$, and then it applies a non-linear activation function $g^{[l]}$ to this linear combination. It can be written as follow:
        \begin{equation}
            A^{[l]} = g^{[l]}(W^{[l]}A^{[l-1]}+b^{[l]})\text{,}
            \label{eq:nn}
        \end{equation}
        where $W^{[l]}$ and $b^{[l]}$ are the weights and the biases of the layer $l$, respectively, that need to be learned. 
        
        The activation function, denoted by $g^{[l]}$ in Equation~\ref{eq:nn}, is crucial as it allows to introduce non-linear combinations of the features. If only linear functions are used, the depth of the network will have little effect since the final output will simply be a linear combination of the input, which could be achieved with only one layer. In this work, we use the well-kown Rectified Linear Units (ReLU), calculated as $\text{ReLU}(z) = \max(0,z)$ \cite{krizhevsky_2012}.
        

  
        Stacking several layers allows to increase the capacity of the network to represent complex functions, while keeping the layers simple, \ie composed of a small number of units. Section~\ref{subsec:complexity_res} will provide some experimental results for different network depths. 
        
        Let~$(X,Y)$ be a set of $n$ training instances such as $(X,Y)=\{(\mathbf{x_1},y_1),(\mathbf{x_2},y_2),\cdots,(\mathbf{x_n},y_n) \}\in \mathbb{R}^{T\times D}\times \mathcal{Y}$. The pair $(\mathbf{x_i},y_i)$ represents training instance $i$ where $\mathbf{x_i}$ is a~$D$-variate time series of length $T$ associated with the label $y_i\in \mathcal{Y}=\{1,\cdots,C\}$ for $C$ classes. Formally, $\mathbf{x_i}$ can be expressed by $\mathbf{x_i}=\langle x_i(1),\cdots,x_i(T) \rangle$, where $x_i(t)=\left(x_i^1(t),\cdots,x_i^D(t)\right)$ for a time stamp $t$. Note that $A^{[0]}$ is equal to $X$ in Equation~\ref{eq:nn}.

        Training a neural network corresponds to finding the values of $\mathbf{W}=\{W^{[l]}\}_{\forall l}$ and $\mathbf{b}=\{b^{[l]}\}_{\forall l}$ that will minimize a given cost function, which assesses the fit of the model to the data. This process is known as empirical risk minimization, and the cost function $\mathcal{J}$ is usually defined as the average of the errors committed on each training instance:
        \begin{equation}
            \mathcal{J}(\mathbf{W},\mathbf{b}) = \frac{1}{n}\sum_{\mathbf{x_i}}\mathcal{L}(\hat{y_i},y_i)\text{.}
            \label{eq:cost}
        \end{equation}
        where $\hat{y_i}$ correspond to the network predictions.
        
        The loss function $\mathcal{L}(\hat{y_i},y_i)$ is usually expressed for a multi-class problem as the cross-entropy loss:
        \begin{align}
            \mathcal{L}(\hat{y_i},y_i) &= -\sum_{y \in \{ 1, \cdots , C\} } \mathbbm{1}\{ y=y_{i}\}log(p(y|x_i))\\
                        &= -log(p(y_i|x_i))\text{,}
            \label{eq:loss}
        \end{align}
        where $p(y_i|x_i)$ represents the probability of predicting the true class $y_i$ of instance $i$ computed by the last layer of the network, the Softmax layer, and denoted by $A^{[L]}$ for a network composed of $L$ layers.. 
        
        
        
        Training deep neural networks presents two main challenges which are offset by a substantial benefit. First, it requires significant expertise to engineer the architecture of the network, choose its hyper-parameters, and decide how to optimize it. In return, such models require less feature engineering than more traditional classification algorithms and have shown to provide superior accuracy across a wide range of tasks. It is in some sense shifting the difficulty of engineering the features to the one of engineering the architecture. Second, deep neural networks are usually prone to overfitting because of their very low bias: they have so many parameters that they can fit a very large family of distributions, which in turn creates an overfitting issue \cite{zhang_2016}. Section~\ref{subsec:complexity_res} will provide an analysis of TempCNN accuracy as a function of the number of learned parameters.
        
        
     \subsection{Temporal Convolutions}
        
        Convolutional layers were proposed to limit the number of weights that a network needs to learn while trying to make the most of the structuring dimensions in the data --~\eg{}spatial, temporal or spectral~-- \cite{lecun_1990}. They apply a convolution filter to the output of the previous layer. Conversely to the dense layer (\ie the fully-connected layer presented in Section~\ref{subsec:intro_cnn}) where the output of a neuron is a single number reflecting the activation, the output of a convolutional layer is therefore a set of activations. For example, if the input is a univariate time series, then the output will be a time series where each point in the series is the result of a convolution filter.
        
\begin{figure}[!ht]
            \centering
            \includegraphics[width=0.55\linewidth]{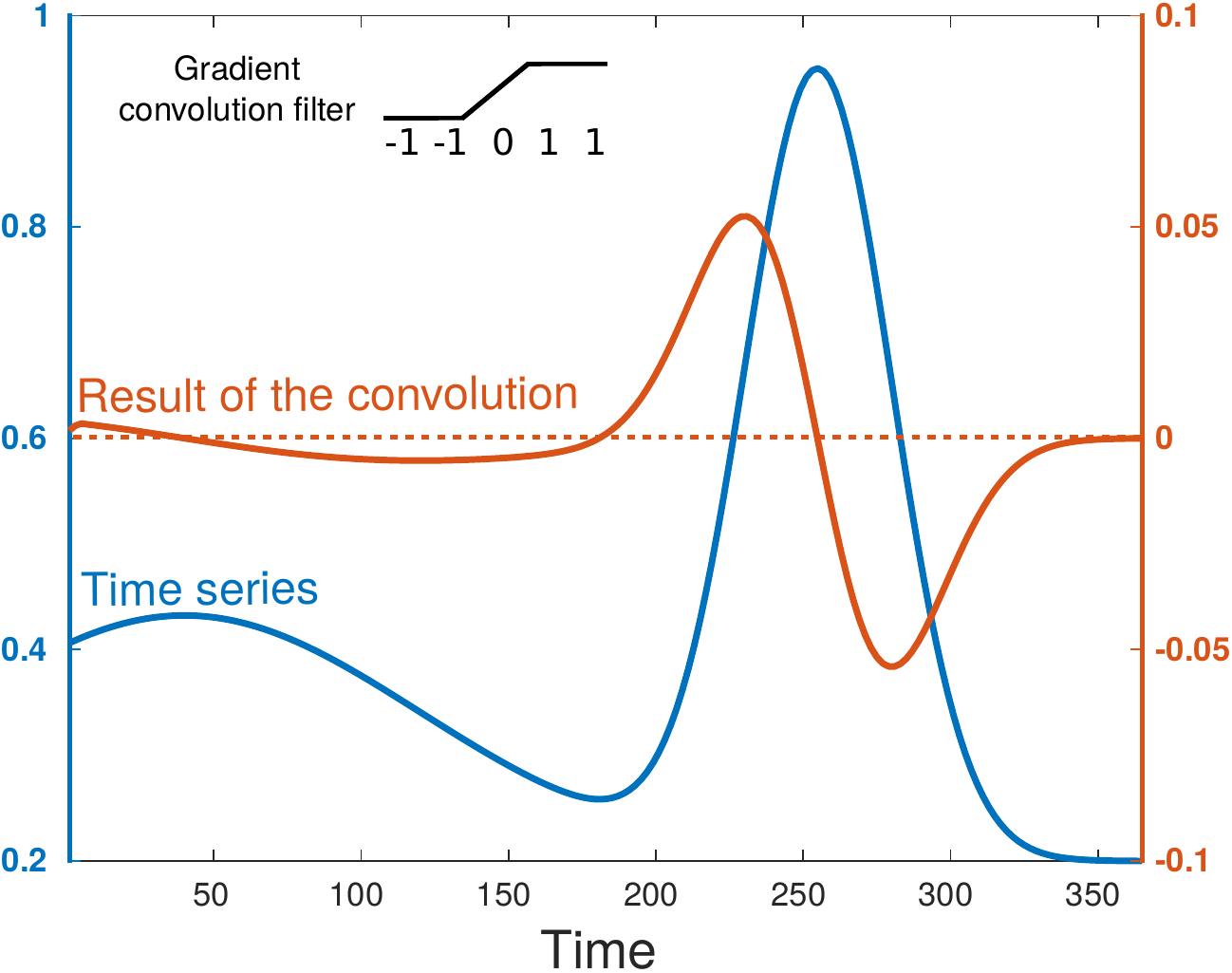}
            \caption{Convolution of a time series (blue) with the positive gradient filter $[-1\ -1\ 0\ 1\ 1]$ (black). The result (red) takes high positive values when the signal sharply increases, and conversely.}
            \label{fig:example_conv}
        \end{figure}

        Figure~\ref{fig:example_conv} shows the application of a gradient filter $[-1\ -1\ 0\ 1\ 1]$ onto the time series depicted in blue. The output is depicted in red. It takes high positive values where an increase in the signal is detected, and low negative values where a decrease in the signal occurs. Note that the so-called convolution is technically a cross-correlation.
        
        Compared to dense layers that apply different weights to the different inputs, convolutional layers differ in that they share their parameters: the same linear combination is applied by sliding it over the entire input. This drastically reduces the number of weights in the layer, by assuming that the same convolution might be useful in different parts of the time series. This is why the number of trainable parameters only depends on the filter size of the convolution $f$ and on the number of units $n$, but not on the size of the input. Conversely, the size of the output will depend on the size of the input, and also on two other hyper-parameters --~the stride and the padding. The stride represents the interval between two convolution centers. The padding controls for the size after the application of the convolution by adding values (usually zeros) at the borders of the input.

    \subsection{Proposed network architecture}
    \label{subsec:our_temporalCNN}

    In this section, we present the baseline architecture that will be discussed in this manuscript. The goal is not to propose the best architecture through exhaustive experiments, but rather to explore the behavior of TempCNNs for SITS classification. 
    
\begin{figure*}[ht]
        	\centering
        	\includegraphics[width=\linewidth]{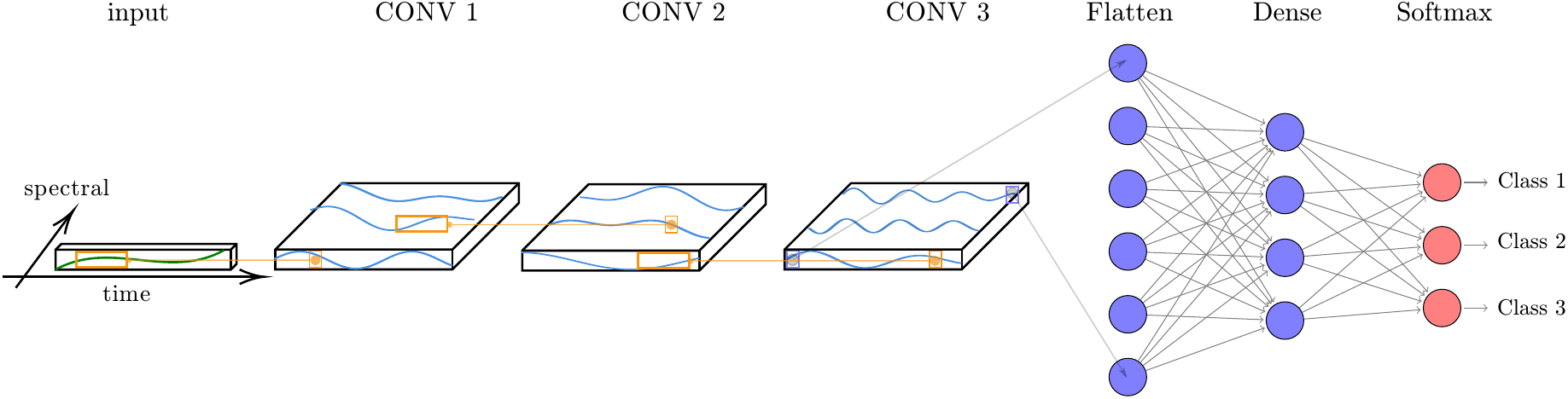}	
            \caption{Proposed temporal Convolutional Neural Network (TempCNN). The network input is a multi-variate time series. Three convolutional filters are consecutively applied, then one dense layer, and finally the Softmax layer, that provides the predicting class distribution.}
        	\label{fig:our_architecture}
        \end{figure*}

    Figure~\ref{fig:our_architecture} displays the TempCNN architecture used in the experiments (Section~\ref{sec:exp_results}). The architecture is composed of three convolutional layers (64 units), one dense layer (256 units) and one Softmax layer. The filter size of convolutions is set to 5. Section~\ref{subsec:complexity_res} will justify the width (\ie number of units) and the depth (\ie number of convolution layers) of this architecture. Moreover, Section \ref{subsec:filter_size} will study the influence of convolution filter size. Finally, the experimental section will also examine the use of pooling layers \ref{subsec:pooling_res}.

    To control for overfitting issues, we use four regularization mechanisms: batch normalisation \cite{ioffe_2015}, dropout with a dropout rate of 0.5 \cite{srivastava_2014}, a $\mathcal{L}_2$-regularization on the weights (also named weight-decay) applied for all the layers with a rate of $10^{-6}$, and  validation set, corresponding to 5~\% of the training set.
    Similarly to the split between the train and test set (see the details in Section~\ref{subsec:ref_data}), the validation set is composed of instances that cannot come from the same polygons of the train set. Section~\ref{subsec:controll_overfitting_res} will detail the influence of these four regularization mechanisms.
    
	
    In the experiments, the parameters of the studied networks are trained using Adam optimization (standard parameter values: $\beta_1=0.9$, $\beta_2=0.999$, and  $\epsilon=10^{-8}$) \cite{kingma_2014} for a batch size equal to 32, and a number of epochs equal to 20. An early stopping mechanism with a patience of zero is also applied. The infleunce of the batch size on the accuracy and the training time would be analyzed in Section~\ref{subsec:batch_size_res}.
    
    
    All the studied CNN models have been implemented with Keras library \cite{chollet_2015}, with Tensorflow as the backend \cite{abadi_2016}. To facilitate others to build on this work, we have made our code available at \codeRepo.

\section{Material and Methods}
\label{sec:data}

	This section presents the dataset used for the experiments. First, optical satellite data are presented. Next, the used reference data are briefly described. Then, the data preparation steps are detailed. Finally, benchmark algorithms and evaluation measure are presented.

    \subsection{Optical Satellite Data}

\begin{figure}[ht]
	        \centering
            \includegraphics[width=0.55\linewidth]{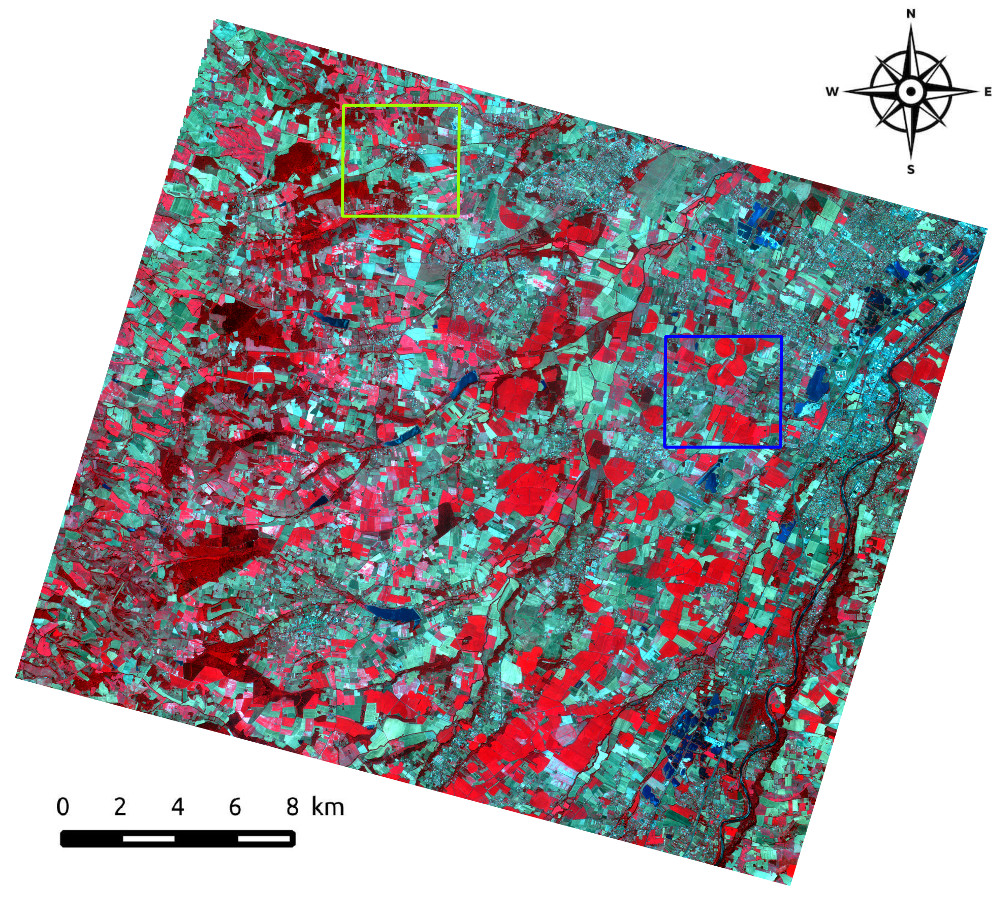}
            \caption{Formosat-2 image in false color for July 14 2006. Green and blue areas will be visually inspected for the experiments.}
            \label{fig:img_area}
        \end{figure}

        The study area is located at the South West of France, near Toulouse city (1\textdegree 10'E, 43\textdegree 27'N). It is 24~km~$\times$~24~km area where about 60~\% of the soil correspond to arable surfaces. The area has a temperate continental climate with hot and dry summer --~average temperature about 22.4~\textdegree C and rainfall about 38~mm per month. Figure~\ref{fig:img_area} displays a satellite image of the area in false color for July 14 2006.

\begin{figure}[ht]
            \centering
        	\includegraphics[width=0.65\linewidth]{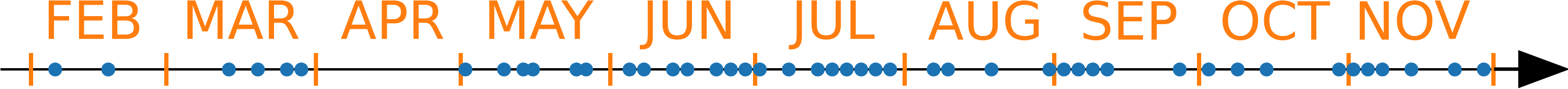}
        	\caption{Acquisition dates of the Formosat-2 image time series.}
        	\label{fig:Formosat_temporal_resolution}
        \end{figure}

        The satellite dataset is composed of 46 Formosat-2 images acquired at 8 meter spatial resolution during the year 2006. Figure~\ref{fig:Formosat_temporal_resolution} shows the distribution of the acquisitions, that are mainly concentrated during the summer time. Note that Formosat-2's characteristics are similar to the new Sentinel-2 satellites that provide 10 meter spatial resolution images every five days.
	
        For each Formosat-2 image, only the three bands Near-Infrared (760-900 nm) (NIR), Red (630-690 nm) (R) and Green (520-600 nm) (G) are used. The blue channel has been discarded since it is sensitive to atmospheric artifacts. 

        Each image has been ortho-rectified to ensure the same pixel location throughout the whole time series. In addition, the digital numbers from the row images have been converted to top-of-canopy reflectance by the French Space Agency \cite{hagolle_2015}. This last step corrects images from atmospheric effects, and also outputs cloud, shadow and saturation masks. The remaining steps of the data preparation -- temporal sampling, feature extraction and feature normalization -- are presented in Section~\ref{subsec:data_preparation}.

    \subsection{Reference Data}
    \label{subsec:ref_data}

	    The reference data come from three sources: 1) farmer's declaration from 2006 (\textit{Registre Parcellaire Graphique} in French), 2) ground field campaigns performed during 2006, and 3) a reference map obtained with a semi-automatic procedure \cite{idbraim_2009}. From these three reference sources, a total of 13 classes is extracted representing three winter crops (wheat, barley and rapeseed), five summer crops (\eg corn, soy and sunflower), four natural classes (grassland, forests and water) and the urban surfaces. Note that the reference map is used only to extract the urban surfaces, and each extracted urban polygon is visually controlled.

        Table~\ref{tab:number_instances} displays the total number of instances per class at pixel- and polygon-level. It shows great variations in the number of available instances for each class where grassland, urban surfaces, wheat and sunflower predominate.
    
        \begin{table}[ht]
    	    \caption{Number of instances per class counted at pixel- and polygon-level.}
            \label{tab:number_instances}
            \centering
            \begin{tabular}{lrrc}
        	    \hline
        	    \textbf{Classes} & \textbf{Pixels} & \textbf{Polygons} & \textbf{Legend} \\
                \hline
                \textbf{Wheat} & 194,699 & 295 & \raisebox{.4\totalheight}{\fcolorbox{black}{wheat}{}} \\
                \textbf{Barley} & 23,404 & 43 &  \raisebox{.4\totalheight}{\fcolorbox{black}{barley}{}} \\
                \textbf{Rapeseed} & 36,720 & 55 & \raisebox{.4\totalheight}{\fcolorbox{black}{rapeseed}{}}\\
                \textbf{Corn} & 62,885 & 83 & \raisebox{.4\totalheight}{\fcolorbox{black}{corn}{}} \\
                \textbf{Soy} & 9481 & 24 & \raisebox{.4\totalheight}{\fcolorbox{black}{soy}{}} \\
                \textbf{Sunflower} & 108,718 & 173 & \raisebox{.4\totalheight}{\fcolorbox{black}{sunflower}{}} \\
                \textbf{Sorghum} & 17,305 & 22 & \raisebox{.4\totalheight}{\fcolorbox{black}{sorghum}{}} \\
                \textbf{Pea} & 9151 & 15 & \raisebox{.4\totalheight}{\fcolorbox{black}{pea}{}} \\
                \textbf{Grassland} & 202,718 & 328 & \raisebox{.4\totalheight}{\fcolorbox{black}{grassland}{}} \\
                \textbf{Deciduous} & 29,488 & 24 & \raisebox{.4\totalheight}{\fcolorbox{black}{deciduous}{}}\\
                \textbf{Conifer} & 15,818 & 18 & \raisebox{.4\totalheight}{\fcolorbox{black}{conifer}{}} \\
                \textbf{Water} & 30,544 & 32 & \raisebox{.4\totalheight}{\fcolorbox{black}{water}{}}\\
                \textbf{Urban} & 292,478 & 307 & \raisebox{.4\totalheight}{\fcolorbox{black}{urban}{}} \\
                \hline
                \textbf{Total} & 1,033,409 & 1419 & \\
                \hline
            \end{tabular}
        \end{table}
    
        The reference data are randomly split into two independent datasets at the polygon level where 60 \% of the data is used for training the classification algorithms and 40 \% is used for testing.  To statistically evaluate the performance of the different algorithms, this split operation is repeated five times. Hence, each algorithm is evaluated five times on different train/test splits. The presented results are averaged over these five folds. 
        
        \subsection{Data Preparation}
    	    \label{subsec:data_preparation}

		    \subsubsection{Temporal sampling}
            \label{subsubsec:temporal_sampling_strategy}
        	
            The optical SITS includes invalid pixels due to the presence of clouds and saturated pixels. Nowadays, the high temporal resolution of SITS is used to efficiently detect clouds and their shadows \cite{hagolle_2015}. The produced masks are then used to gap-filled the cloudy and saturated pixels before applying supervised classification algorithms without a loss of accuracy \cite{inglada_2015}. 
            We use here a temporal linear interpolation for imputing invalid pixel values. 

            As most of the classification algorithms explored in the manuscript require a regular temporal sampling, we apply  interpolation on a regular temporal grid defined with a time gap of two days. The starting and ending dates correspond to the first and last acquisition dates of the Formosat-2 series, respectively. This operation artificially increases the length of the Formosat-2 time series from 46 to 149. As some studied algorithms, such as RF, may be sensitive to this increase of the length, the temporal interpolation is also applied for the original sampling. 
            
            \subsubsection{Feature extraction}
            \label{subsubsec:feature_extraction}
        
            Taking benefit from Formosat-2 spectral resolution, spectral indexes are computed after the gapfilling, for each image of Formosat-2 time series. Spectral indexes are commonly used in addition of spectral bands as the input of the supervised classification system in remote sensing literature \cite{gomez_2016}. They can help the classifier to handle some non-linear relationships between the spectral bands \cite{inglada_2017}
            
            More specifically, we compute three commonly used indexes: the Normalized Difference Vegetation Index (NDVI) \cite{rouse_1974}, the Normalized Difference Water Index (NDWI) \cite{mcfeeters_1996} and a brilliance index (IB)  defined as the norm of all the available bands \cite{petitjean_2012, inglada_2015}.
            
            In the experiments, we want to quantify the contribution of the spectral features for the proposed TempCNN models. To this end, a total of three different feature vectors are defined: 1) NDVI only, 2) spectral bands (SB), and 3) SB + NDVI + NDWI + IB. The simplest strategy corresponds to use all the available spectral bands. The contribution of the spectral features is analyzed by adding the three computed spectral indexes (NDVI, NDWI and IB) to the spectral bands. We also decide to analyze separately NDVI, as it is the most common index for vegetation mapping. Table~\ref{tab:number_variables} summarizes the total number of variables for the studied datasets as a function of the temporal strategy and the used spectral features.

            \begin{table}[ht]
                \caption{Number of features for the studied dataset.}
                \label{tab:number_variables}
                \centering
                \begin{tabular}{rccc}              		
                    \hline
                    \textbf{Temporal sampling} & \textbf{NDVI} & \textbf{SB} & \textbf{SB-NDVI-NDWI-IB} \\
                    \hline
                    \textbf{original} & 46 & 138 & 276 \\
                    \textbf{2 days} & 149 & 447 & 894 \\
                    \hline
                \end{tabular}
                ~\\
                \begin{flushleft}
				    {SB: Spectral Bands - NDVI: Normalized Difference Vegetation Index -
            	    NDWI: Normalized Difference Water Index - IB: Brilliance Index}
                \end{flushleft}
            \end{table}
            
     	\subsubsection{Feature normalization}
         
            In remote sensing, the input time series are generally standardized by subtracting the mean and divided by the standard deviation for each feature where each time stamp is considered as a separate feature. This standardization, also called feature scaling, assures that the measure distance, often an Euclidean distance computed through all features, is not dominated by a single feature that has a high dynamic rank. However, it transforms the general temporal trend of the instances.
                        
            In machine learning, the input data are generally $z$-normalized by subtracting the mean and divided by the standard deviation for each time series \cite{bagnall_2017}. This $z$-normalization has been introduced to be able to compare time series that have similar trends, but different scaling and shifting \cite{goldin_1995}. However, it leads to a loss of the significance of the magnitude that it is recognized as crucial for vegetation mapping, \eg the corn will have higher NDVI values than other summer crops.
            
            To overcome both limitations of the common normalization methods, we decide to use a min-max normalization per type of feature. The traditional min-max normalization performs a subtraction of the minimum, then a division by the range, \ie the maximum minus the minimum \cite{han_2011}. As this normalization is highly sensitive to extreme values, we propose to use 2~\% (or 98~\%) percentile rather than the minimum (or the maximum) value. For each feature, both percentile values are extracted from all the time-stamp values.
        
        \subsection{Benchmark Algorithms}
        \label{subsec:benchmark_algorithms}    
        
            For the sake of a comparison, we use Random Forests (RF) as a traditional state-of-the-art algorithm for SITS classification, as well as Recurrent Neural Networks (RNNs) as deep learning leading algorithm. Both benchmark algorithms are briefly introduced.
            
            \subsubsection{Random Forests}
            The remote sensing community has assessed the performance of different algorithms for SITS classification showing that Random Forests (RF) and the Support Vector Machines (SVM) algorithms dominate the other traditional algorithms \cite{belgiu_2016,gomez_2016}. In particular, the RF algorithm manages the high dimension of the SITS data \cite{pelletier_2016}, is robust to the presence of mislabeled data \cite{pelletier_2017}, has high accuracy performance on large scale study \cite{inglada_2017}, and has parameters easy to tune \cite{pelletier_2016}.       
            
            The RF algorithm builds an ensemble of binary decision trees \cite{breiman_2001}. Its first specificity is to use bootstrap instances at each tree --~\ie training instances randomly selected with replacement \cite{breiman_1996}~-- to increase the diversity among the trees. The second specificity is the use of random subspace technique for choosing the splitting criterion at each node: a subset of the features is first randomly selected, then all the possible splits on this subset are tested based on a feature value test, \eg maximization of the Gini index. It will result in a split of the data into two subsets, for which previous operations are recursively repeated. The construction stops when all the nodes are pure (\ie in each node, all the data belong to the same class), or when a user-defined criterion is met, such as a maximum depth or a number of instances at the node below a threshold.

            To complete the experiments of Section \ref{sec:exp_results}, the RF implementation from Scikit-Learn has been used with standard parameter settings \cite{pelletier_2016}: 500 trees at the maximum depth, and a number of randomly selected variables per node equals to the square root of the total number of features.
            
            \subsubsection{Recurrent Neural Networks}
            
            First developed for sequential data, RNN models have been recently applied to several classification tasks in remote sensing, especially for crop mapping \cite{minh_2018, ndikumana_2018,russwurm_2018}. They share the learned features across different positions. As the error is back-propagated at each time step, the computational cost can be high and it might cause learning issues such as vanishing gradient. Hence, most recent RNN architectures use LSTM or GRU units that help to capture long distance connections and solve the vanishing gradient issue. They are composed of a memory cell as well as update and reset gates to decide how much new information to add and how much past information to forget.
            
            Following the most recent studies, we have trained bidirectional RNNs composed of the stack of three GRUs, one dense layer (256 neurons) and a Softmax layer \cite{russwurm_2018,ndikumana_2018}.
            The same number of neurons is used in the three GRUs. More specifically, we have trained five different models comprised of 16, 32, 64, 128 or 256 neurons. As we focus on the analysis of TempCNNs in this paper, Section~\ref{subsec:spectral_temporal_help_res} will report only the best RNN result (128 neurons) obtained on the test instances. All the models have been trained similarly to CNNs with Keras (backend Tensorflow): batch size equal to 32, Adam optimization, and monitoring of the validation loss with an early stopping mechanism (\textit{c.f.} Section~\ref{subsec:our_temporalCNN}).

            
    \subsection{Performance Evaluation}
    
    	The performance of the different classification algorithms are quantitatively and qualitatively evaluated. Following traditional quantitative evaluations, confusion matrices are obtained by comparing the referenced labels with the predicted ones. Then, the standard  Overall Accuracy (OA) measure is computed. In addition, the results will be also qualitatively evaluated through a visual inspection.

\section{Experimental Results}
\label{sec:exp_results}
    
	This section aims at evaluating the TempCNN architecture presented in Section~\ref{subsec:our_temporalCNN}. A set of six experiments is run to study:
    \begin{enumerate}
    	\item how the proposed CNN models benefit from both spectral and temporal dimensions,
    	\item how the filter size of temporal convolutions influence the performance,
    	\item how pooling layers influence the performance,
    	\item how big and deep the model should be,
    	\item how the regularization mechanisms help the learning,
    	\item what values used for the batch size.  
    \end{enumerate}
    A last section is dedicated to the visual analysis of the produced land cover maps.

    As explained in Section~\ref{subsec:ref_data}, all the presented Overall Accuracy (OA) values correspond to average values over five folds. When displayed, the interval always correspond to one standard deviation. Moreover, one can see all the details of the trained networks at \codeRepo.

      \subsection{Benefiting from both spectral and temporal dimensions.}
      \label{subsec:spectral_temporal_help_res}
      	
      
      
        In this experiment, we compare four configurations: 1) no guidance, 2) only temporal guidance, 3) only spectral guidance, and 4) both temporal and spectral guidance. Before presenting the obtained results, we first describe the trained models for these fourth types of guidance.
      
        \textbf{No guidance}: Similar to a traditional classifier, such as the RF algorithm, the first considered type of model ignores the spectral and temporal structures of the data, \ie a shuffle of the data across both spectral and temporal dimensions will provide the same results. For this configuration, we decided to train two types of algorithms: 1) the RF classifier selected as the competitor \cite{gomez_2016}, and 2) a deep learning model composed of three dense layers of 1024 units --~ this specific architecture is named FC in the following. As both RF and FC models do not require regular temporal sampling, the use of 2-day sampling is not necessary and can even lead to under-performance. Indeed, the use of high dimensional space composed of redundant and sometimes noisy features may hurt the accuracy performance. Hence, the results of both models are displayed for the original temporal sampling. 
        
        \textbf{Temporal guidance}: The second type of model provides guidance only on the temporal dimension. Among all the possible architectures, we decided to train an architecture with convolution filters of size $(f,1)$, instead of $(f,D)$ with $D$ the number of features. In other words, same convolution filters are applied across the temporal dimension identically for all the spectral dimensions. 

        \textbf{Spectral guidance:} The third type of model includes guidance only on the spectral dimension. For this purpose, a convolution of size $(1,D)$ is first applied without padding, reducing the spectral dimension to one for the next convolution layers of size $(1,1)$.
        
        \textbf{Spectro-temporal guidance:} The last type of model corresponds to the one presented in Section~\ref{subsec:our_temporalCNN}, where the first convolution has a size $(f,D)$. The choice of this architecture is explained in the following sections. For the sake of comparison, RNNs that provide also spectro-temporal guidance (Section~\ref{subsec:benchmark_algorithms}) are also evaluated.

		Table~\ref{tab:domain_help} displays the Overall Accuracy (OA) values and one standard deviation for the four levels of guidance. As the use of engineering features may help the different models, we train all the models for the three feature vectors presented in Section~\ref{subsec:data_preparation}: NDVI alone, spectral bands (SB), and spectral bands with three spectral indexes (SB-SF). For both models using temporal guidance, the filter size $f$ is set to five. All the models are learned as specified in Section~\ref{subsec:our_temporalCNN}, including dropout and batch normalization layers, weight decay and the use of a validation set.

        \begin{table}[ht]
            \centering
            \caption{Averaged overall accuracy ($\pm$ one standard deviation) over five folds for four levels of help in spectral and/or temporal dimensions. Three feature vectors are used here: 1) Normalized Difference vegetation index (NDVI), 2) spectral bands (SB), and 3) SB and three spectral features (SB-SF). Bold values show the highest performance for each type of features. CNN/ COnvolutional Neural Network}
            \label{tab:domain_help}
    		\begin{tabular}{llccc}
            \hline
            	& & \textbf{NDVI} & \textbf{SB} & \textbf{SB-SF}\\
            \hline
           \textbf{No guidance} & \textbf{Random Forest (RF)} & 88.17$\pm$0.59  & 90.02$\pm$1.44 & 90.92$\pm$1.22 \\
            
             & \textbf{Fully-Connected (FC)} & 86.90$\pm$1.65 & 91.36$\pm$1.15  & 91.87$\pm$0.88\\
            
            \hline
            \multicolumn{2}{l}{\textbf{Recurrent Neural Network (RNN)}}  & 88.62$\pm$0.86 & 92.18$\pm$1.43 & 92.38$\pm$0.83\\
            \hline
            \multicolumn{2}{l}{\textbf{CNN with temporal guidance}} & \textbf{90.16$\pm$0.94} & 92.74$\pm$0.80 & 93.00$\pm$0.83 \\
            \hdashline
            \multicolumn{2}{l}{\textbf{\textcolor{white}{CNN} with spectral guidance}} & 88.24$\pm$0.63& 93.34$\pm$0.88& 93.24$\pm$0.83\\
            \hdashline
            \multicolumn{2}{l}{ \textbf{ {\textcolor{white}{CNN}}with spectro-temporal guidance (TempCNN)}} & 90.06$\pm$0.88 & \textbf{93.42$\pm$0.76} & \textbf{93.45$\pm$0.77} \\
            \hline
            \end{tabular}
    	\end{table}
        
       Table~\ref{tab:domain_help} shows that CNN OA increases when adding more types of guidance, and that regardless of the type of used features. Note that the case of using only spectral guidance with NDVI feature is a particular ``degenerate'' case: the spectral dimension is composed of only one feature (NDVI). The trained model applies convolutions of size (1,1), leading to a model that does not provide with guidance.

       When using at least the spectral bands in the feature vector (SB and SB-SF columns), RNN, FC and RF models obtain lower accuracy with higher variations than TempCNNs: OA differences varies from 1~to~3~\%. Interestingly, models based on only spectral convolutions with spectral features (fifth row, second column) slightly outperform models that used only temporal guidance (fourth row, second column). This result confirms thus the importance of the spectral domain for land cover mapping application. In addition, the use of convolutions in both temporal and spectral domains leads to slightly better OA compared to the other three levels of guidance. Finally, Table~\ref{tab:domain_help} shows that the use of spectral indexes in addition of the available spectral bands does not help to improve the accuracy of traditional and deep learning algorithms.

    \subsection{Influence of the filter size}
    \label{subsec:filter_size}
    
        For CNN models using a temporal guidance, it is also interesting to study the filter size. Considering the 2-day regular temporal sampling, a filter size of $f$ (with $f$ an odd number) will abstract the temporal information over $\pm (f-1)$ days, before and after each point of the series. Given this natural expression in number of days, we name $(f-1)$ the reach of the convolution: it corresponds to half of the width of the temporal neighborhood used for the temporal convolutions. 

        Table~\ref{tab:temporal_filter_help} displays the OA values as a function of reach for TempCNN. We study five size of filters $f=\{3,5,9,17,33\}$ corresponding to a reach of 2, 4, 8, 16, and 32 days, respectively. 
        
        \begin{table}[ht!]
            \centering
            \caption{Averaged Overall Accuracy (OA), $\pm$ one standard deviation, for five reach values. Bold value highlights the highest OA value.}
            \label{tab:temporal_filter_help}
            \begin{tabular}{rccccc}
                \hline
                \textbf{Reach} & \textbf{2} & \textbf{4} & \textbf{8} & \textbf{16} & \textbf{32} \\
                \hline
                \textbf{OA} & 93.28$\pm$0.82 & 93.42$\pm$0.76 & \textbf{93.43$\pm$0.62} & 93.00$\pm$0.85 & 92.79$\pm$0.72 \\
                \hline
            \end{tabular}
        \end{table}
        
                
        
        Table~\ref{tab:temporal_filter_help} shows the maximum OA is reached for a reach of 8 days, with a similiar OA for 4 days.  This result shows the importance of high temporal resolution SITS, such as the one provided at five days by both Sentinel-2 satellites. Indeed, the acquisition frequency will allow CNNs to abstract enough temporal information from the temporal convolutions. In general, the reach of the convolutions will mainly depend on the patterns that need to be abstracted at a given temporal resolution.
        


    \subsection{Are local and global temporal pooling layers important?}
    \label{subsec:pooling_res}
    
        In this Section, we explore the use of pooling layers for different reach values. Pooling layers are generally used in image classification task to both speed-up the computation and make more robust to noise some of the learned features \cite{boureau_2010}. They can be seen as a de-zooming operation, which naturally induces a multi-scale analysis when interleaved between successive convolutional layers. For a time series, these pooling layers simply reduce the length, and thus the resolution, of the time series that are output by the neurons --~and this by a factor $k$. 
        
        Two types of pooling have received most of the attention in the literature about image analysis: 1)~the local max-pooling \cite{ren_2015}, and 2) the global average pooling \cite{he_2016}. For time series, global average pooling seems to have been more successful  \cite{wang_2017,fawaz_2018}. We want to see here if these previous results can be generalized for time series classification.
        
    
        For this purpose, we train models with a global average pooling layer added after the third convolution layers for the following reach: 2, 4, 8, 16, and 32 days. We also train models with local pooling layers interleaved between each convolution layer with a window size $k$ of 2. For this experiment, the reach is kept constant --~2, 4, 8, 16, and 32 days~-- by reducing the convolution filter size $f$ after each convolution. For example, a constant reach of 8 is obtained by applying successively three convolutions with filter sizes of 9, 5, and 3. 
    
\begin{figure}[ht]
       	    \centering 
		    \includegraphics[width=0.45\linewidth]{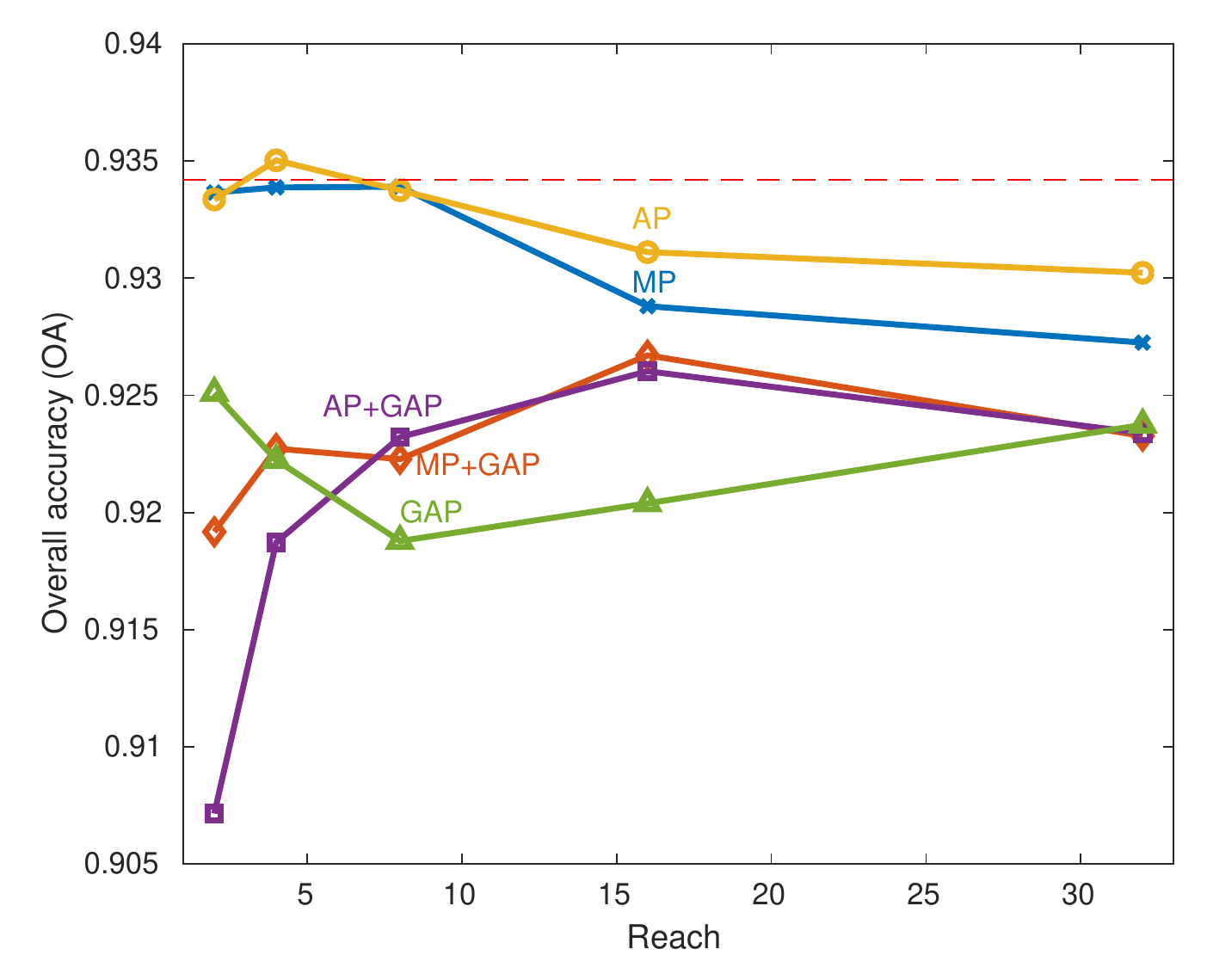}
    	    \caption{Overall Accuracy (OA) as function of reach for local max-pooling (MP) in blue, local max-pooling and global average pooling (MP+GAP) in orange, local average pooling (AP) in yellow, local and global average pooling (AP+GAP) in purple, and global average pooling (GAP) alone in green. The used dataset is composed of the three spectral bands with a regular temporal sampling at two days.}
    	    \label{fig:pooling}
    	\end{figure}

        Figure~\ref{fig:pooling} displays the OA values as a function of reach. Each curve represents a different configuration: local max-pooling (MP) in blue, local max-pooling and global average pooling (MP+GAP) in orange, local average pooling (AP) in yellow, local and global average pooling (AP+GAP) in purple, and global average pooling (GAP) in green. The horizontal red dashed line corresponds to the OA values obtained without pooling layers in the previous experiment. 
     
        Figure~\ref{fig:pooling} shows that the use of pooling layers performs poorly: the OA results are almost always below the one obtained without pooling layers (red dashed line). Let us describe in more details the different findings for both global and local pooling layers.
     
        The use of a global average pooling layer leads to the biggest decrease in accuracy. This layer is generally used to drastically reduce the number of trainable parameters by reducing the size of the last convolution layer to its depth. It thus performs an extreme dimensionality reduction, that decreases here the accuracy performance. 
     
        Regarding the use of only local pooling layers, Figure~\ref{fig:pooling} shows similar results for both max and average pooling layers. The OA values tend to decrease when the reach increases. The results are similar to those obtained by the model without pooling layers (horizontal red dashed line) for reach values lower than nine days, with even a slight improvement when using a local average pooling layer with a constant reach of four days.
        
        This last result is in disagreement with results obtained for image classification tasks for which: 1) max-pooling tends to have better results than average pooling, and 2) the use of local pooling layers help to improve the classification performance. The main reason for this difference is probably task-related. In image classification, local pooling layers are known to extract features that are invariant to the scale and small transformations leading to models that can detect objects in an image no matter their locations or their sizes. However, the location of the temporal features, and their amplitude, are crucial for SITS classification. For example, winter and summer crops, that we want to distinguish, may have similar profiles with only a shift in time. Removing the temporal location of the peak of greenness might prevent their discrimination.
     
    
    \subsection{How big and deep model for our data?}
    \label{subsec:complexity_res}
        
        \subsubsection{Width influence or the bias-variance trade-off}
        
        This section has two goals: 1)~justify the complexity of the learned models, and 2)~be able to give an idea about how big should the model be.
        Both objectives relate to the bias-variance trade-off of the model for our quantity of data. The more complex the model (\ie more parameters), the lower its bias, \ie{}the fewer incorrect assumptions the model makes about the distribution from which the data is sampled. Conversely, given a fixed quantity of training data, the more complex the model, the higher the variance. Many classifiers vary their bias-variance trade-off automatically, such as when decision trees grow deeper as the quantity of data increases. For neural networks however, the bias is fixed by the architecture, and the variance for a fixed quantity of data.  We use here the number of trainable parameters as a proxy for model complexity, which provides a reasonable measure when dealing with a specific classification problem where the quantity of data and the number of classes are fixed \cite[Chapter 7, Introduction]{goodfellow_2016}.

\begin{figure}[ht]
      	    \centering
            \begin{subfloat}[Width influence]{\includegraphics[height=5.6cm]{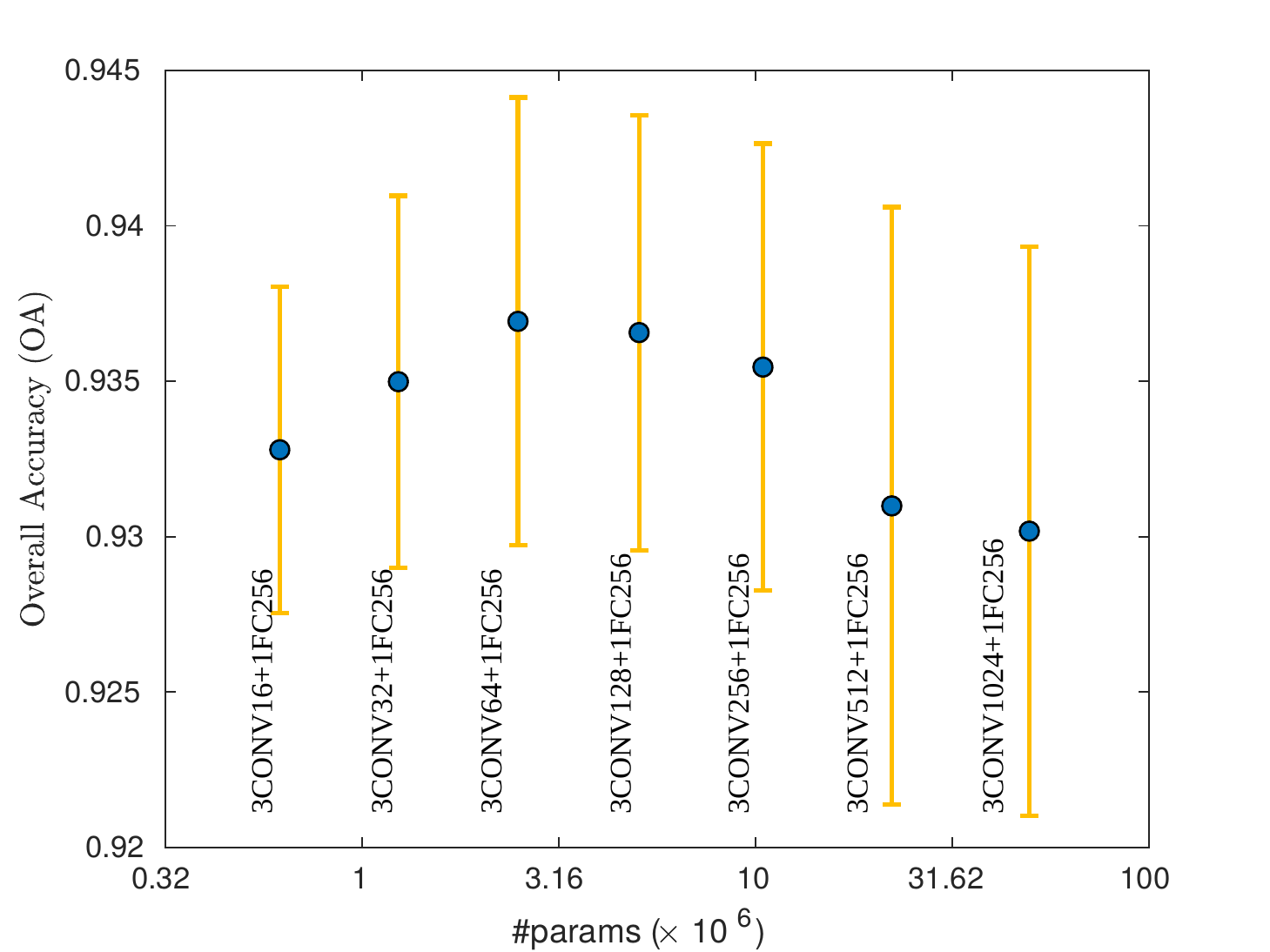}\label{subfig:width}}
            \end{subfloat}
           \begin{subfloat}[Depth influence]{\includegraphics[height=5.2cm]{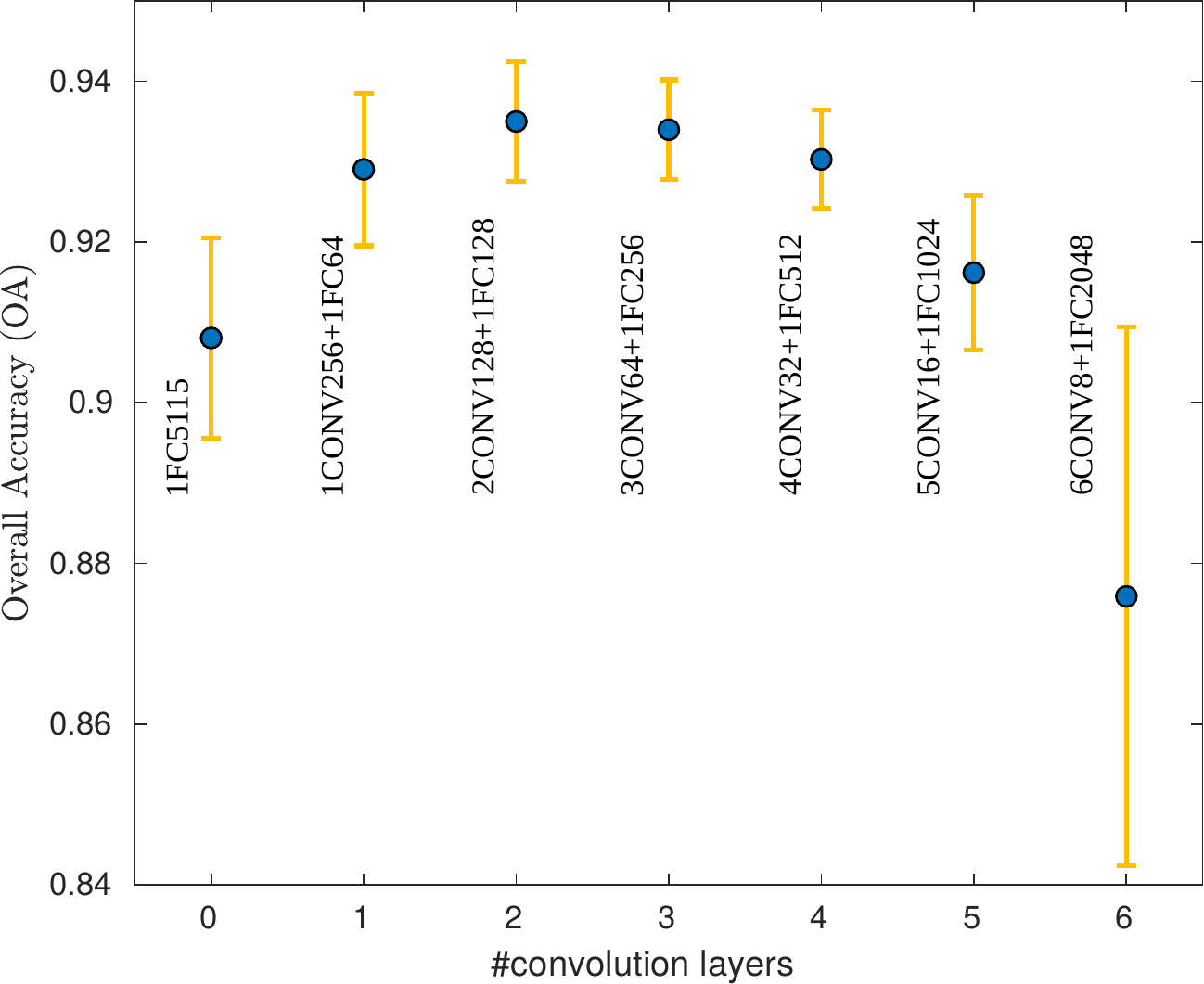}\label{subfig:depth}}\end{subfloat}
            \caption{Overall Accuracy ($\pm$ one standard deviation) as a function of the number of parameters (width influence) or the number of convolution networks (depth influence) for seven temporal Convolutional Neural Network models. The orange bar represents one standard deviation. The used dataset is composed of three spectral bands with a regular temporal sampling at two days.}
            \label{fig:width_depth}
        \end{figure}

        We studied seven CNN architectures with increasing number of parameters. Each architecture is composed of three convolutional layers, one dense layer (256 neruons), and the Softmax layer as depicted in Figure~\ref{fig:our_architecture}. We then vary the number of neurons --~or \emph{width}~--of the convolutional layers (16 to 1024 neurons). The depth of the model is specifically studied right after.
        The total number of trainable parameters then ranges from about 320,000 to 50 million. All models are trained following Section~\ref{subsec:our_temporalCNN}, with the exception of not using a validation set in order to observe more accuracy variations by letting the models being more prone to overfitting. Figure~\ref{subfig:width} shows the OA values as a function of number of parameters in logarithmic-scale.
        

       
        Figure~\ref{subfig:width} shows that the architecture is very robust to a drastic change in number of parameters, as exhibited by OA varying only between 93.28~\% for 50M parameters and 93.69~\% for 2.5M parameters. The standard deviation increases with the number of parameters, but even at 50M parameters\footnote{Note that 50M parameters is an extremely large number of parameters for our dataset of about 620,000 training instances.}, most results are between 91~\% and 95~\% accuracy. From this result, one can see that models having about 2.5M of parameters with three convolutional layers and one dense layers have a good bias-variance trade-off.
       
        If more training data is available, using a similar architecture is likely to conservatively work well, because more data is likely to drive the variance down while bias is fixed. If less data is available, one could decide to use a smaller architecture but again, overall, the results are very stable.
     
    \subsubsection{Depth influence}
    
    	We now propose here to vary the number of layers, \ie{} the depth of the network, for the same model complexities. For this purpose, we decrease the number of units to deeper networks. More specifically, we consider six architectures composed of one to six convolutional layers with a number of units ranging from 256 to 16, and one dense layer with a number of units ranging from 64 to 2048. Figure~\ref{subfig:depth} shows OA values as a function of number of convolutions. 
             
      
        Figure~\ref{subfig:depth} shows that the highest accuracy scores are obtained with the lowest standard deviation for an optimal number of convolutional layers of two or three. For a given complexity (here about 2.5M parameters), the use of an inappropriate number of convolutional layers and number of units may lead to an under-estimation of TempCNNs. The selection of a reasonable architecture might be crucial, and could be thus optimized through computationally expensive cross-validation procedure or meta-learning approaches \cite{bergstra_2012,hutter_2011,snoek_2012}.
    
    \subsection{How to control overfitting?}
    \label{subsec:controll_overfitting_res}
    
        The used TempCNN model includes four mechanisms to deal with overfitting issue: 1)~regularization of the weighs, 2)~dropout, 3)~validation set and 4)~batch normalization layer. Regarding the optimal architecture of Section~\ref{subsec:complexity_res}, the model needs to learn a number of parameters higher than more than three times the given number of training instances~--~2.5M of parameters \textit{versus} 620,000 training data instances. This Section aims at determining which of the used regularization mechanisms are the most crucial to train the TempCNN network. For this purpose, we first trained TempCNN architectures with only one regularization technique, then with all regularization techniques except one.
        
             
		Table~\ref{tab:controll_overfitting} displays OA values with or without the use of different regularization mechanisms. The first row displays the results when no regularization mechanism is applied (lower bound), whereas the last row displays the results when all the regularization mechanisms are used (higher bound).
          
        \begin{table}[ht]
            \centering
            \caption{Averaged overall accuracy over five runs using different regularization techniques. The first and last rows display OA when all the regularization techniques are turned off and on, respectively. The used dataset is composed of three spectral bands with a regular temporal sampling at two days.}
            \label{tab:controll_overfitting}
    		\begin{tabular}{rc}
            \hline
            	& \textbf{Overall Accuracy} \\
            \hline
            \textbf{Nothing} & 90.83 $\pm$ 0.82 \\
            \hline
            \textbf{Only dropout} & 93.12 $\pm$ 0.64 \\
            \textbf{Only batch normalization} & 92.22 $\pm$ 0.86 \\
            \textbf{Only validation set} & 91.17 $\pm$  0.94\\
            \textbf{Only weight decay} & 90.74 $\pm$  1.00 \\  
            \hline
            \textbf{All except dropout} & 92.07 $\pm$ 1.20\\
            \textbf{All except batch normalization} &  92.89 $\pm$ 0.72\\
            \textbf{All except validation set} & 93.68 $\pm$ 0.60\\
            \textbf{All except weight decay} & 93.52 $\pm$ 0.77 \\  
            \hline
            \textbf{All} & 93.42 $\pm$ 0.76\\
            \hline
            \end{tabular}
    	\end{table}
        
        Table~\ref{tab:controll_overfitting} shows that the use of dropout is the most important regularization mechanism for TempCNNs as its only use leads to an OA value close from the one obtained when using the four regularization mechanisms. Conversely, the use of a validation set and the weight decay seems less useful for regularizing the network.

    \subsection{What values are used for the batch size?}
    \label{subsec:batch_size_res}
    	
    	This section aims at studying the influence of the batch size on the classification performance and on the training time. For this purpose, the model of Section~\ref{subsec:controll_overfitting_res} is trained for the following batch sizes: $\{8,16,32,64,128\}$. Table~\ref{tab:batch_size} displays the OA values  and also the training time for each studied batch size.

        \begin{table}[ht!]
        	\centering
            \caption{Training time and averaged Overall Accuracy (OA) over five folds for the same model learned for five batch sizes.}
            \label{tab:batch_size}
        	\begin{tabular}{rrc}
            \hline
            \textbf{Batch size}	& \textbf{Training time} & \textbf{OA}\\
            \hline
            \textbf{8} & 3h45min & 93.54 $\pm$ 0.67\\
            \textbf{16} & 1h56min & 93.65 $\pm$ 0.73\\
            \textbf{32} & 1h06min & 93.59 $\pm$ 0.74\\
            \textbf{64} & 34min & 93.43 $\pm$ 0.71\\
            \textbf{128} & 19min & 93.45 $\pm$ 0.83\\
            \hline
            \end{tabular}
       	\end{table}
        
        Table~\ref{tab:batch_size} shows that in the case of our experiments the batch size influences the training time, but not the accuracy of the learned models, as all OA values are comparable. This result implies that large batch sizes can be selected, if memory storage is available, to speed up the training.
    
    \subsection{Visual analysis}
        
        This experimental section ends with a visual analysis of the results for both blue and green areas of size 3.7~km~$\times$~3.6~km (465 pixels $\times$ 450 pixels) displayed in Figure~\ref{fig:img_area}. The analysis is performed for RF and TempCNN. The original temporal sampling is used for RF, whereas the regular temporal sampling at two days is used to train TempCNN. Both models are learned on datasets composed of the three spectral bands.
            
\begin{figure}[!ht]
    		\includegraphics[width=\linewidth]{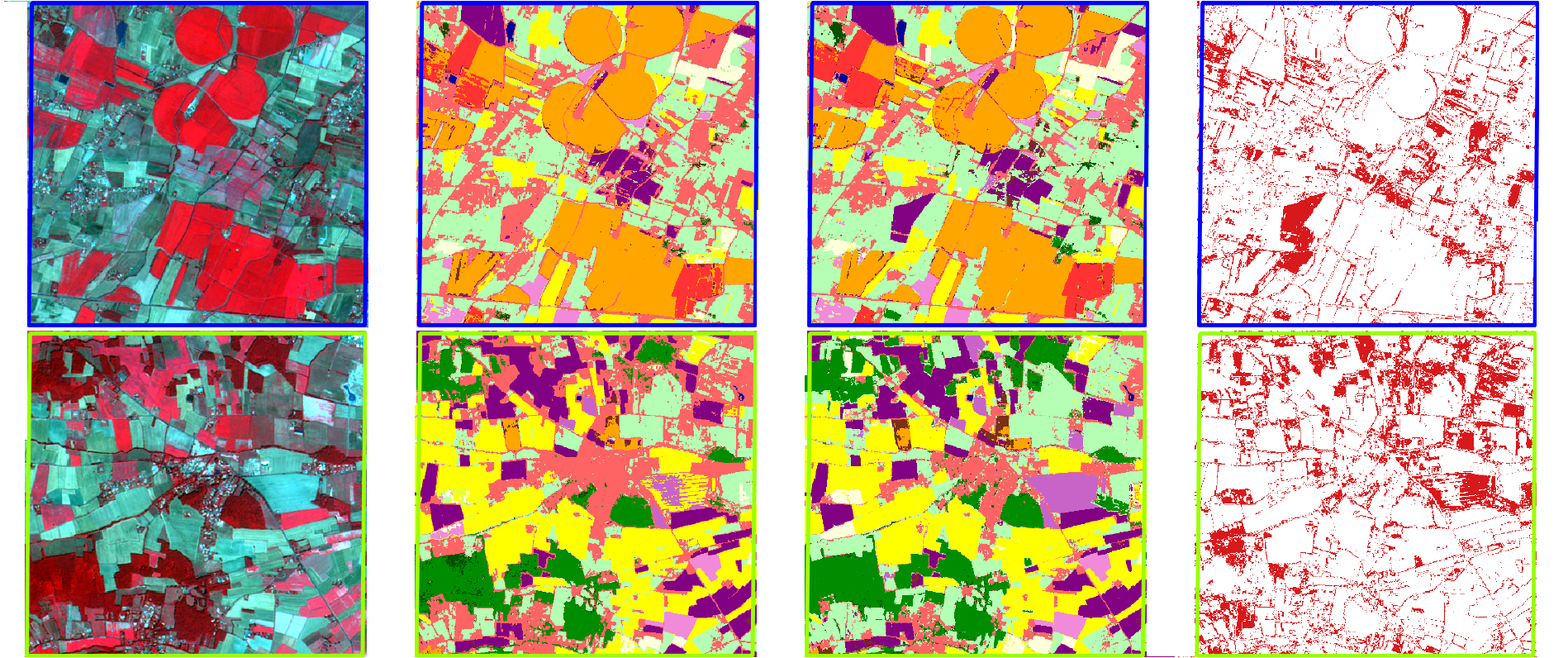}
    		
    		\hspace{0.4cm} Formosat-2 image \hspace{2cm} RF \hspace{2.7cm} TempCNN \hspace{1.7cm} RF $\neq$ TempCNN
        	\caption{Visual results for two areas. The first column displays the Formosat-2 image in false color for July 14 2006. The second and third columns give the results for the Random Forest (RF) and the temporal Convolutional Neural Network (TempCNN) algorithms, respectively. The images in the last column displays in red the disagreement between both classifiers.}
    	    \label{fig:img_visual_analysis}
        \end{figure}

        Figure~\ref{fig:img_visual_analysis} displays the produced land cover maps. The first row displays the results for the blue area, whereas the second row displays the one for the green area. The first column displays the Formosat-2 image in false color for July 14 2006 (zoom of Figure~\ref{fig:Formosat_temporal_resolution}). The second and third columns give the results for the RF and the TempCNN algorithms, respectively. Images in the last column display in red the disagreements between both classifiers. Legend of land cover maps has been provided in Table~\ref{tab:number_instances}.
            
        Although the results look visually similar, the disagreement images between both classifiers highlight some strong differences on the delineations between several land cover, but also at the object-level (\eg crop, urban areas, forest). Regarding the delineation disagreements, we found that RF spreads out the majority class, \ie urban areas, leading to an over-detection of this class, especially for mixing pixels. Regarding object disagreements, one can observe that the RF confounds an urban area (in light pink) that should be, according to the reference (test polygon in that case), a sunflower crop (in purple). Finally, this visual analysis shows that both classification algorithms are sensitive to salt and pepper noise, that could be potentially removed by a post-processing procedure or by incorporating into the classification framework some spatial information.       

\section{Conclusion}
\label{sec:clc}

    This work explores the use of TempCNNs for SITS classification. Through an extensive set of experiments carried out on a series of 46 Formosat-2 images, we show that the tested TempCNN architectures outperform RF and RNN algorithms by 1~to~3~\%. A visual analysis also shows the good quality of temporal CNNs to accurately map land cover without over representation of majority classes.

    To provide intuitions beyond this good performance, we studied the impact of the network architecture by varying the depth and the width of the models, by looking at the influence of common regularization mechanisms, and by testing different batch size values. We have also demonstrated the importance of using both temporal and spectral dimensions when computing the convolutions. The remaining experimental results support two main recommendations on the use of pooling layers and the engineering of spectral features. First, we show that the use of global pooling layers, which drastically reduces the number of trainable parameters, is armful for SITS classification. Overall, we recommend careful study of the influence of pooling layers before any integration into a TempCNN network, and to favor local average global pooling. Second, we show that the addition of manually-calculated spectral features, such as the NDVI, does not seem to improve TempCNN models. We thus recommend not to compute and use them.

    All these results show that TempCNNs are a strong learner for Sentinel-2 SITS, which presents a high spectral resolution with 10 bands at a spatial resolution of 10 and 20 meters. The presence of salt and pepper noise also indicates a need to take into account the spatial dimension of SITS in addition to the spectral and temporal dimensions.
\section*{Acknowledgments}
Dr~Fran\c{c}ois \textsc{Petitjean} is the recipient of an Australian Research Council Discovery Early Career Award (project number DE170100037) funded by the Australian Government.
This material is based upon work supported by the Air Force Office of Scientific Research, Asian Office of Aerospace Research and Development (AOARD) under award number FA2386-18-1-4030.

The authors would like to thank their colleagues from the CESBIO laboratory (Jordi Inglada, Danielle Ducrot, Claire Marais-Sicre, Olivier Hagolle and Mireille Huc) for providing the reference land-cover maps as well as the corrected FORMOSAT-2 images. They would also like to thank Hassan Ismail Fawaz and Germain Forestier for their comments on the manuscript. Future work should therefore propose framework to take into account the spatial dimension of SITS.

\authorcontributions{All the authors have been involved in the writing of the manuscript and the design of experiments. Dr.~Charlotte \textsc{Pelletier} has run the different experiments.
Investigation, Charlotte \textsc{Pelletier}; Methodology, Charlotte \textsc{Pelletier}, Geoffrey I. \textsc{Webb} and Fran\c{c}ois \textsc{Petitjean}; Writing – review \& editing, Charlotte \textsc{Pelletier}, Geoffrey I. \textsc{Webb} and Fran\c{c}ois \textsc{Petitjean}.}

\conflictofinterests{The authors declare no conflict of interest.} 

\abbreviations{The following abbreviations are used in this manuscript:\\

\noindent AP: Average Pooling\\
CNN: Convolutional Neural Network\\
GAP: Global Average Pooling\\
IB: Brilliance Index\\
MP: Max pooling\\
NDVI: Normalized Difference Vegetation Index\\
NDWI: Normalized Difference Water Index\\
OA: Overall Accuracy\\
RF: Random Forests\\
RNN: Recurrent Neural Network\\
SB: Spectral Band\\
SITS: Satellite Image Time Series\\
SVM: Support Vector machine\\
TempCNN: Temporal Convolutional Neural Network}



\bibliographystyle{mdpi}

\renewcommand\bibname{References}

\bibliography{biblio.bib}

\begin{thebibliography}{-------}
\providecommand{\natexlab}[1]{#1}

\bibitem[Bojinski \em{et~al.}(2014)Bojinski, Verstraete, Peterson, Richter,
  Simmons, and Zemp]{bojinski_2014}
Bojinski, S.; Verstraete, M.; Peterson, T.C.; Richter, C.; Simmons, A.; Zemp,
  M.
\newblock The concept of essential climate variables in support of climate
  research, applications, and policy.
\newblock {\em Bulletin of the American Meteorological Society} {\bf 2014},
  {\em 95},~1431--1443.

\bibitem[Feddema \em{et~al.}(2005)Feddema, Oleson, Bonan, Mearns, Buja, Meehl,
  and Washington]{feddema_2005}
Feddema, J.J.; Oleson, K.W.; Bonan, G.B.; Mearns, L.O.; Buja, L.E.; Meehl,
  G.A.; Washington, W.M.
\newblock The importance of land-cover change in simulating future climates.
\newblock {\em Science} {\bf 2005}, {\em 310},~1674--1678.

\bibitem[G{\'o}mez \em{et~al.}(2016)G{\'o}mez, White, and Wulder]{gomez_2016}
G{\'o}mez, C.; White, J.C.; Wulder, M.A.
\newblock Optical remotely sensed time series data for land cover
  classification: A review.
\newblock {\em ISPRS Journal of Photogrammetry and Remote Sensing} {\bf 2016},
  {\em 116},~55--72.

\bibitem[Inglada \em{et~al.}(2017)Inglada, Vincent, Arias, Tardy, Morin, and
  Rodes]{inglada_2017}
Inglada, J.; Vincent, A.; Arias, M.; Tardy, B.; Morin, D.; Rodes, I.
\newblock Operational high resolution land cover map production at the country
  scale using satellite image time series.
\newblock {\em Remote Sensing} {\bf 2017}, {\em 9},~95.

\bibitem[Drusch \em{et~al.}(2012)Drusch, Del~Bello, Carlier, Colin, Fernandez,
  Gascon, Hoersch, Isola, Laberinti, Martimort, Meygret, Spoto, Sy, Marchese,
  and Bargellini]{drusch_2012}
Drusch, M.; Del~Bello, U.; Carlier, S.; Colin, O.; Fernandez, V.; Gascon, F.;
  Hoersch, B.; Isola, C.; Laberinti, P.; Martimort, P.; Meygret, A.; Spoto, F.;
  Sy, O.; Marchese, F.; Bargellini, P.
\newblock {Sentinel-2: ESA's} Optical High-Resolution Mission for {GMES}
  Operational Services.
\newblock {\em Remote Sensing of Environment} {\bf 2012}, {\em 120},~25--36.

\bibitem[Matton \em{et~al.}(2015)Matton, Sepulcre, Waldner, Valero, Morin,
  Inglada, Arias, Bontemps, Koetz, and Defourny]{matton_2015}
Matton, N.; Sepulcre, G.; Waldner, F.; Valero, S.; Morin, D.; Inglada, J.;
  Arias, M.; Bontemps, S.; Koetz, B.; Defourny, P.
\newblock An automated method for annual cropland mapping along the season for
  various globally-distributed agrosystems using high spatial and temporal
  resolution time series.
\newblock {\em Remote Sensing} {\bf 2015}, {\em 7},~13208--13232.

\bibitem[Vuolo \em{et~al.}(2018)Vuolo, Neuwirth, Immitzer, Atzberger, and
  Ng]{vuolo_2018}
Vuolo, F.; Neuwirth, M.; Immitzer, M.; Atzberger, C.; Ng, W.T.
\newblock How much does multi-temporal {Sentinel-2} data improve crop type
  classification?
\newblock {\em International Journal of Applied Earth Observation and
  Geoinformation} {\bf 2018}, {\em 72},~122--130.

\bibitem[Immitzer \em{et~al.}(2016)Immitzer, Vuolo, and
  Atzberger]{immitzer_2016}
Immitzer, M.; Vuolo, F.; Atzberger, C.
\newblock First experience with {Sentinel-2} data for crop and tree species
  classifications in central {Europe}.
\newblock {\em Remote Sensing} {\bf 2016}, {\em 8},~166.

\bibitem[Ismail~Fawaz \em{et~al.}(2018)Ismail~Fawaz, Forestier, Weber,
  Idoumghar, and Muller]{fawaz_2018}
Ismail~Fawaz, H.; Forestier, G.; Weber, J.; Idoumghar, L.; Muller, P.A.
\newblock Deep learning for time series classification: a review.
\newblock {\em arXiv preprint arXiv:1809.04356} {\bf 2018}.

\bibitem[Bengio \em{et~al.}(2013)Bengio, Courville, and Vincent]{bengio_2013}
Bengio, Y.; Courville, A.; Vincent, P.
\newblock Representation learning: A review and new perspectives.
\newblock {\em IEEE Transactions on Pattern Analysis and Machine Intelligence}
  {\bf 2013}, {\em 35},~1798--1828.

\bibitem[Zhu \em{et~al.}(2017)Zhu, Tuia, Mou, Xia, Zhang, Xu, and
  Fraundorfer]{zhu_2017}
Zhu, X.X.; Tuia, D.; Mou, L.; Xia, G.S.; Zhang, L.; Xu, F.; Fraundorfer, F.
\newblock Deep Learning in Remote Sensing: A Comprehensive Review and List of
  Resources.
\newblock {\em IEEE Geoscience and Remote Sensing Magazine} {\bf 2017}, {\em
  5},~8--36.

\bibitem[Khatami \em{et~al.}(2016)Khatami, Mountrakis, and
  Stehman]{khatami_2016}
Khatami, R.; Mountrakis, G.; Stehman, S.V.
\newblock A meta-analysis of remote sensing research on supervised pixel-based
  land-cover image classification processes: General guidelines for
  practitioners and future research.
\newblock {\em Remote Sensing of Environment} {\bf 2016}, {\em 177},~89--100.

\bibitem[Jia \em{et~al.}(2014)Jia, Liang, Wei, Yao, Su, Jiang, and
  Wang]{jia_2014}
Jia, K.; Liang, S.; Wei, X.; Yao, Y.; Su, Y.; Jiang, B.; Wang, X.
\newblock {Land cover classification of Landsat data with phenological features
  extracted from time series MODIS NDVI data}.
\newblock {\em Remote Sensing} {\bf 2014}, {\em 6},~11518--11532.

\bibitem[Pittman \em{et~al.}(2010)Pittman, Hansen, Becker-Reshef, Potapov, and
  Justice]{pittman_2010}
Pittman, K.; Hansen, M.C.; Becker-Reshef, I.; Potapov, P.V.; Justice, C.O.
\newblock Estimating global cropland extent with multi-year {MODIS} data.
\newblock {\em Remote Sensing} {\bf 2010}, {\em 2},~1844--1863.

\bibitem[Valero \em{et~al.}(2016)Valero, Morin, Inglada, Sepulcre, Arias,
  Hagolle, Dedieu, Bontemps, Defourny, and Koetz]{valero_2016}
Valero, S.; Morin, D.; Inglada, J.; Sepulcre, G.; Arias, M.; Hagolle, O.;
  Dedieu, G.; Bontemps, S.; Defourny, P.; Koetz, B.
\newblock Production of a dynamic cropland mask by processing remote sensing
  image series at high temporal and spatial resolutions.
\newblock {\em Remote Sensing} {\bf 2016}, {\em 8},~55.

\bibitem[Pelletier \em{et~al.}(2016)Pelletier, Valero, Inglada, Champion, and
  Dedieu]{pelletier_2016}
Pelletier, C.; Valero, S.; Inglada, J.; Champion, N.; Dedieu, G.
\newblock Assessing the robustness of {Random Forests} to map land cover with
  high resolution satellite image time series over large areas.
\newblock {\em Remote Sensing of Environment} {\bf 2016}, {\em 187},~156--168.

\bibitem[Petitjean \em{et~al.}(2012)Petitjean, Inglada, and
  Gan{\c{c}}arski]{petitjean_2012}
Petitjean, F.; Inglada, J.; Gan{\c{c}}arski, P.
\newblock Satellite image time series analysis under time warping.
\newblock {\em IEEE Transactions on Geoscience and Remote Sensing} {\bf 2012},
  {\em 50},~3081--3095.

\bibitem[Maus \em{et~al.}(2016)Maus, C{\^a}mara, Cartaxo, Sanchez, Ramos, and
  de~Queiroz]{maus_2016}
Maus, V.; C{\^a}mara, G.; Cartaxo, R.; Sanchez, A.; Ramos, F.M.; de~Queiroz,
  G.R.
\newblock A time-weighted dynamic time warping method for land-use and
  land-cover mapping.
\newblock {\em IEEE Journal of Selected Topics in Applied Earth Observations
  and Remote Sensing} {\bf 2016}, {\em 9},~3729--3739.

\bibitem[Belgiu and Csillik(2018)]{belgiu_2018}
Belgiu, M.; Csillik, O.
\newblock Sentinel-2 cropland mapping using pixel-based and object-based
  time-weighted dynamic time warping analysis.
\newblock {\em Remote Sensing of Environment} {\bf 2018}, {\em 204},~509--523.

\bibitem[Schroff \em{et~al.}(2015)Schroff, Kalenichenko, and
  Philbin]{schroff_2015}
Schroff, F.; Kalenichenko, D.; Philbin, J.
\newblock Facenet: A unified embedding for face recognition and clustering.
\newblock  Proceedings of the IEEE Conference on Computer Vision and Pattern
  Recognition,  2015, pp. 815--823.

\bibitem[Redmon and Farhadi(2017)]{redmon_2017}
Redmon, J.; Farhadi, A.
\newblock {YOLO9000}: Better, Faster, Stronger.
\newblock  2017 IEEE Conference on Computer Vision and Pattern Recognition
  (CVPR). IEEE,  2017, pp. 6517--6525.

\bibitem[Bahdanau \em{et~al.}(2014)Bahdanau, Cho, and Bengio]{bahdanau_2014}
Bahdanau, D.; Cho, K.; Bengio, Y.
\newblock Neural machine translation by jointly learning to align and
  translate.
\newblock  International Conference on Learning Representations (ICLR),  2014.

\bibitem[Krizhevsky \em{et~al.}(2012)Krizhevsky, Sutskever, and
  Hinton]{krizhevsky_2012}
Krizhevsky, A.; Sutskever, I.; Hinton, G.E.
\newblock Imagenet classification with deep convolutional neural networks.
\newblock  Advances in Neural Information Processing Systems,  2012, pp.
  1097--1105.

\bibitem[Ioffe and Szegedy(2015)]{ioffe_2015}
Ioffe, S.; Szegedy, C.
\newblock Batch Normalization: {Accelerating} Deep Network Training by Reducing
  Internal Covariate Shift.
\newblock  International Conference on Machine Learning,  2015, pp. 448--456.

\bibitem[Maggiori \em{et~al.}(2017)Maggiori, Tarabalka, Charpiat, and
  Alliez]{maggiori_2017}
Maggiori, E.; Tarabalka, Y.; Charpiat, G.; Alliez, P.
\newblock {Convolutional Neural Networks} for large-scale remote-sensing image
  classification.
\newblock {\em IEEE Transactions on Geoscience and Remote Sensing} {\bf 2017},
  {\em 55},~645--657.

\bibitem[Postadjian \em{et~al.}(2017)Postadjian, Le~Bris, Sahbi, and
  Mallet]{postadjian_2017}
Postadjian, T.; Le~Bris, A.; Sahbi, H.; Mallet, C.
\newblock Investigating the potential of deep neural networks for large-scale
  classification of very high resolution satellite images.
\newblock  ISPRS Annals of the Photogrammetry Remote Sensing and Spatial
  Information Sciences,  2017, Vol.~4.

\bibitem[Volpi and Tuia(2017)]{volpi_2017}
Volpi, M.; Tuia, D.
\newblock Dense semantic labeling of subdecimeter resolution images with
  convolutional neural networks.
\newblock {\em IEEE Transactions on Geoscience and Remote Sensing} {\bf 2017},
  {\em 55},~881--893.

\bibitem[Audebert \em{et~al.}(2017)Audebert, Le~Saux, and
  Lef{\`e}vre]{audebert_2017}
Audebert, N.; Le~Saux, B.; Lef{\`e}vre, S.
\newblock Segment-before-detect: Vehicle detection and classification through
  semantic segmentation of aerial images.
\newblock {\em Remote Sensing} {\bf 2017}, {\em 9},~368.

\bibitem[Zhang \em{et~al.}(2018)Zhang, Yuan, Zeng, Li, and Wei]{zhang_2018}
Zhang, Q.; Yuan, Q.; Zeng, C.; Li, X.; Wei, Y.
\newblock Missing Data Reconstruction in Remote Sensing image with a Unified
  Spatial-Temporal-Spectral Deep Convolutional Neural Network.
\newblock {\em IEEE Transactions on Geoscience and Remote Sensing} {\bf 2018},
  {\em 56},~4274--4288.

\bibitem[Masi \em{et~al.}(2016)Masi, Cozzolino, Verdoliva, and
  Scarpa]{masi_2016}
Masi, G.; Cozzolino, D.; Verdoliva, L.; Scarpa, G.
\newblock {Pansharpening by Convolutional Neural Networks}.
\newblock {\em Remote Sensing} {\bf 2016}, {\em 8},~594.

\bibitem[Liang and Li(2016)]{liang_2016}
Liang, H.; Li, Q.
\newblock Hyperspectral imagery classification using sparse representations of
  convolutional neural network features.
\newblock {\em Remote Sensing} {\bf 2016}, {\em 8},~99.

\bibitem[Hu \em{et~al.}(2015)Hu, Xia, Hu, and Zhang]{hu_2015}
Hu, F.; Xia, G.S.; Hu, J.; Zhang, L.
\newblock Transferring deep convolutional neural networks for the scene
  classification of high-resolution remote sensing imagery.
\newblock {\em Remote Sensing} {\bf 2015}, {\em 7},~14680--14707.

\bibitem[Li \em{et~al.}(2017)Li, Zhang, and Shen]{li_2017}
Li, Y.; Zhang, H.; Shen, Q.
\newblock Spectral--spatial classification of hyperspectral imagery with {3D}
  convolutional neural network.
\newblock {\em Remote Sensing} {\bf 2017}, {\em 9},~67.

\bibitem[Hamida \em{et~al.}(2018)Hamida, Benoit, Lambert, and
  Amar]{hamida_2018}
Hamida, A.B.; Benoit, A.; Lambert, P.; Amar, C.B.
\newblock {3-D} Deep Learning Approach for Remote Sensing Image Classification.
\newblock {\em IEEE Transactions on Geoscience and Remote Sensing} {\bf 2018},
  {\em 56},~4420--4434.

\bibitem[Kussul \em{et~al.}(2017)Kussul, Lavreniuk, Skakun, and
  Shelestov]{kussul_2017}
Kussul, N.; Lavreniuk, M.; Skakun, S.; Shelestov, A.
\newblock Deep learning classification of land cover and crop types using
  remote sensing data.
\newblock {\em IEEE Geoscience and Remote Sensing Letters} {\bf 2017}, {\em
  14},~778--782.

\bibitem[Scarpa \em{et~al.}(2018)Scarpa, Gargiulo, Mazza, and
  Gaetano]{scarpa_2018}
Scarpa, G.; Gargiulo, M.; Mazza, A.; Gaetano, R.
\newblock A {CNN}-Based Fusion Method for Feature Extraction from {Sentinel}
  Data.
\newblock {\em Remote Sensing} {\bf 2018}, {\em 10},~236.

\bibitem[Wang \em{et~al.}(2017)Wang, Yan, and Oates]{wang_2017}
Wang, Z.; Yan, W.; Oates, T.
\newblock Time series classification from scratch with deep neural networks: A
  strong baseline.
\newblock  International Joint Conference on Neural Networks (IJCNN). IEEE,
  2017, pp. 1578--1585.

\bibitem[Wu \em{et~al.}(2015)Wu, Wang, Jiang, Ye, and Xue]{wu_2015}
Wu, Z.; Wang, X.; Jiang, Y.G.; Ye, H.; Xue, X.
\newblock Modeling spatial-temporal clues in a hybrid deep learning framework
  for video classification.
\newblock  Proceedings of the 23rd ACM International Conference on Multimedia.
  ACM,  2015, pp. 461--470.

\bibitem[Di~Mauro \em{et~al.}(2017)Di~Mauro, Vergari, Basile, Ventola, and
  Esposito]{dimauro_2017}
Di~Mauro, N.; Vergari, A.; Basile, T.M.A.; Ventola, F.G.; Esposito, F.
\newblock End-to-end Learning of Deep Spatio-temporal Representations for
  Satellite Image Time Series Classification.
\newblock  European Conference on Machine Learning \& Principles and Practice
  of Knowledge Discovery in Databases (PKDD/ECML),  2017.
\newblock Challenge on Time Series Land cover Classification (TiSeLaC).

\bibitem[Zhong \em{et~al.}(2019)Zhong, Hu, and Zhou]{zhong_2019}
Zhong, L.; Hu, L.; Zhou, H.
\newblock Deep learning based multi-temporal crop classification.
\newblock {\em Remote Sensing of Environment} {\bf 2019}, {\em 221},~430--443.

\bibitem[Ji \em{et~al.}(2018)Ji, Zhang, Xu, Shi, and Duan]{ji_2018}
Ji, S.; Zhang, C.; Xu, A.; Shi, Y.; Duan, Y.
\newblock {3D} Convolutional Neural Networks for Crop Classification with
  Multi-Temporal Remote Sensing Images.
\newblock {\em Remote Sensing} {\bf 2018}, {\em 10},~75.

\bibitem[RuBwurm and K{\"o}rner(2017)]{russwurm_2017}
RuBwurm, M.; K{\"o}rner, M.
\newblock Temporal Vegetation Modelling Using Long Short-Term Memory Networks
  for Crop Identification from Medium-Resolution Multi-spectral Satellite
  Images.
\newblock  Computer Vision and Pattern Recognition Workshops,  2017, pp.
  1496--1504.

\bibitem[Sun \em{et~al.}(2018)Sun, Di, and Fang]{sun_2018}
Sun, Z.; Di, L.; Fang, H.
\newblock Using {Long Short-Term Memory Recurrent Neural Network} in land cover
  classification on {Landsat} and Cropland data layer time series.
\newblock {\em International Journal of Remote Sensing} {\bf 2018}, pp. 1--22.

\bibitem[Ienco \em{et~al.}(2017)Ienco, Gaetano, Dupaquier, and
  Maurel]{ienco_2017}
Ienco, D.; Gaetano, R.; Dupaquier, C.; Maurel, P.
\newblock Land Cover Classification via Multitemporal Spatial Data by Deep
  {Recurrent Neural Networks}.
\newblock {\em IEEE Geoscience and Remote Sensing Letters} {\bf 2017}, {\em
  14},~1685--1689.

\bibitem[Minh \em{et~al.}(2018)Minh, Ienco, Gaetano, Lalande, Ndikumana, Osman,
  and Maurel]{minh_2018}
Minh, D.H.T.; Ienco, D.; Gaetano, R.; Lalande, N.; Ndikumana, E.; Osman, F.;
  Maurel, P.
\newblock Deep recurrent neural networks for winter vegetation quality mapping
  via multitemporal {SAR Sentinel-1}.
\newblock {\em IEEE Geoscience and Remote Sensing Letters} {\bf 2018}, {\em
  15},~464--468.

\bibitem[Ndikumana \em{et~al.}(2018)Ndikumana, Ho~Tong~Minh, Baghdadi,
  Courault, and Hossard]{ndikumana_2018}
Ndikumana, E.; Ho~Tong~Minh, D.; Baghdadi, N.; Courault, D.; Hossard, L.
\newblock Deep Recurrent Neural Network for Agricultural Classification using
  multitemporal {SAR Sentinel-1 for Camargue, France}.
\newblock {\em Remote Sensing} {\bf 2018}, {\em 10},~1217.

\bibitem[Benedetti \em{et~al.}(2018)Benedetti, Ienco, Gaetano, Ose, Pensa, and
  Dupuy]{benedetti_2018}
Benedetti, P.; Ienco, D.; Gaetano, R.; Ose, K.; Pensa, R.G.; Dupuy, S.
\newblock M3-Fusion: A Deep Learning Architecture for Multiscale Multimodal
  Multitemporal Satellite Data Fusion.
\newblock {\em IEEE Journal of Selected Topics in Applied Earth Observations
  and Remote Sensing} {\bf 2018}, {\em 11},~4939--4949.

\bibitem[Ru{\ss}wurm and K{\"o}rner(2018)]{russwurm_2018}
Ru{\ss}wurm, M.; K{\"o}rner, M.
\newblock Multi-temporal land cover classification with sequential recurrent
  encoders.
\newblock {\em ISPRS International Journal of Geo-Information} {\bf 2018}, {\em
  7},~129.

\bibitem[Lyu \em{et~al.}(2016)Lyu, Lu, and Mou]{lyu_2016}
Lyu, H.; Lu, H.; Mou, L.
\newblock Learning a transferable change rule from a recurrent neural network
  for land cover change detection.
\newblock {\em Remote Sensing} {\bf 2016}, {\em 8},~506.

\bibitem[Mou \em{et~al.}(2018)Mou, Bruzzone, and Zhu]{mou_2018}
Mou, L.; Bruzzone, L.; Zhu, X.X.
\newblock Learning spectral-spatial-temporal features via a recurrent
  convolutional neural network for change detection in multispectral imagery.
\newblock {\em arXiv preprint arXiv:1803.02642} {\bf 2018}.

\bibitem[Jia \em{et~al.}(2017)Jia, Khandelwal, Nayak, Gerber, Carlson, West,
  and Kumar]{jia_2017}
Jia, X.; Khandelwal, A.; Nayak, G.; Gerber, J.; Carlson, K.; West, P.; Kumar,
  V.
\newblock Incremental dual-memory {LSTM} in land cover prediction.
\newblock  Proceedings of the 23rd ACM SIGKDD International Conference on
  Knowledge Discovery and Data Mining. ACM,  2017, pp. 867--876.

\bibitem[Goodfellow \em{et~al.}(2016)Goodfellow, Bengio, and
  Courville]{goodfellow_2016}
Goodfellow, I.; Bengio, Y.; Courville, A.
\newblock {\em Deep Learning}; MIT Press,  2016.
\newblock \url{http://www.deeplearningbook.org}.

\bibitem[Zhang \em{et~al.}(2017)Zhang, Bengio, Hardt, Recht, and
  Vinyals]{zhang_2016}
Zhang, C.; Bengio, S.; Hardt, M.; Recht, B.; Vinyals, O.
\newblock Understanding deep learning requires rethinking generalization.
\newblock  International Conference on Learning Representations (ICLR),  2017.

\bibitem[LeCun \em{et~al.}(1990)LeCun, Boser, Denker, Henderson, Howard,
  Hubbard, and Jackel]{lecun_1990}
LeCun, Y.; Boser, B.E.; Denker, J.S.; Henderson, D.; Howard, R.E.; Hubbard,
  W.E.; Jackel, L.D.
\newblock Handwritten digit recognition with a back-propagation network.
\newblock  Advances in Neural Information Processing Systems,  1990, pp.
  396--404.

\bibitem[Srivastava \em{et~al.}(2014)Srivastava, Hinton, Krizhevsky, Sutskever,
  and Salakhutdinov]{srivastava_2014}
Srivastava, N.; Hinton, G.; Krizhevsky, A.; Sutskever, I.; Salakhutdinov, R.
\newblock Dropout: a simple way to prevent neural networks from overfitting.
\newblock {\em The Journal of Machine Learning Research} {\bf 2014}, {\em
  15},~1929--1958.

\bibitem[Kingma and Ba(2014)]{kingma_2014}
Kingma, D.P.; Ba, J.
\newblock Adam: A method for stochastic optimization.
\newblock  International Conference on Learning Representations (ICLR),  2014.

\bibitem[Chollet \em{et~al.}(2015)Chollet et~al.]{chollet_2015}
Chollet, F.; others.
\newblock Keras,  2015.
\newblock \url{https://keras.io}.

\bibitem[Abadi \em{et~al.}(2016)Abadi, Barham, Chen, Chen, Davis, Dean, Devin,
  Ghemawat, Irving, Isard, et~al.]{abadi_2016}
Abadi, M.; Barham, P.; Chen, J.; Chen, Z.; Davis, A.; Dean, J.; Devin, M.;
  Ghemawat, S.; Irving, G.; Isard, M.; others.
\newblock TensorFlow: A System for Large-Scale Machine Learning.
\newblock  OSDI,  2016, Vol.~16, pp. 265--283.

\bibitem[Hagolle \em{et~al.}(2015)Hagolle, Huc, Villa~Pascual, and
  Dedieu]{hagolle_2015}
Hagolle, O.; Huc, M.; Villa~Pascual, D.; Dedieu, G.
\newblock A multi-temporal and multi-spectral method to estimate aerosol
  optical thickness over land, for the atmospheric correction of {FormoSat-2,
  LandSat, VENUS and Sentinel-2 images}.
\newblock {\em Remote Sensing} {\bf 2015}, {\em 7},~2668--2691.

\bibitem[Idbraim \em{et~al.}(2009)Idbraim, Ducrot, Mammass, and
  Aboutajdine]{idbraim_2009}
Idbraim, S.; Ducrot, D.; Mammass, D.; Aboutajdine, D.
\newblock An unsupervised classification using a novel {ICM} method with
  constraints for land cover mapping from remote sensing imagery.
\newblock {\em International Review on Computers and Software} {\bf 2009}, {\em
  4},~165--176.

\bibitem[Inglada \em{et~al.}(2015)Inglada, Arias, Tardy, Hagolle, Valero,
  Morin, Dedieu, Sepulcre, Bontemps, Defourny, et~al.]{inglada_2015}
Inglada, J.; Arias, M.; Tardy, B.; Hagolle, O.; Valero, S.; Morin, D.; Dedieu,
  G.; Sepulcre, G.; Bontemps, S.; Defourny, P.; others.
\newblock Assessment of an operational system for crop type map production
  using high temporal and spatial resolution satellite optical imagery.
\newblock {\em Remote Sensing} {\bf 2015}, {\em 7},~12356--12379.

\bibitem[Rouse~Jr \em{et~al.}(1974)Rouse~Jr, Haas, Schell, and
  Deering]{rouse_1974}
Rouse~Jr, J.; Haas, R.; Schell, J.; Deering, D.
\newblock Monitoring vegetation systems in the {Great Plains with ERTS}.
\newblock  Third Symposium on Significant Results Obtained from the first
  Earth,  1974.

\bibitem[McFeeters(1996)]{mcfeeters_1996}
McFeeters, S.K.
\newblock The use of the {Normalized Difference Water Index (NDWI)} in the
  delineation of open water features.
\newblock {\em International journal of remote sensing} {\bf 1996}, {\em
  17},~1425--1432.

\bibitem[Bagnall \em{et~al.}(2017)Bagnall, Lines, Bostrom, Large, and
  Keogh]{bagnall_2017}
Bagnall, A.; Lines, J.; Bostrom, A.; Large, J.; Keogh, E.
\newblock The great time series classification bake off: a review and
  experimental evaluation of recent algorithmic advances.
\newblock {\em Data Mining and Knowledge Discovery} {\bf 2017}, {\em
  31},~606--660.

\bibitem[Goldin and Kanellakis(1995)]{goldin_1995}
Goldin, D.Q.; Kanellakis, P.C.
\newblock On similarity queries for time-series data: constraint specification
  and implementation.
\newblock  International Conference on Principles and Practice of Constraint
  Programming. Springer,  1995, pp. 137--153.

\bibitem[Han \em{et~al.}(2011)Han, Pei, and Kamber]{han_2011}
Han, J.; Pei, J.; Kamber, M.
\newblock {\em Data mining: concepts and techniques}; Elsevier,  2011.

\bibitem[Belgiu and Dr{\u{a}}gu{\c{t}}(2016)]{belgiu_2016}
Belgiu, M.; Dr{\u{a}}gu{\c{t}}, L.
\newblock {Random Forest} in remote sensing: A review of applications and
  future directions.
\newblock {\em ISPRS Journal of Photogrammetry and Remote Sensing} {\bf 2016},
  {\em 114},~24--31.

\bibitem[Pelletier \em{et~al.}(2017)Pelletier, Valero, Inglada, Champion,
  Marais~Sicre, and Dedieu]{pelletier_2017}
Pelletier, C.; Valero, S.; Inglada, J.; Champion, N.; Marais~Sicre, C.; Dedieu,
  G.
\newblock Effect of training class label noise on classification performances
  for land cover mapping with satellite image time series.
\newblock {\em Remote Sensing} {\bf 2017}, {\em 9},~173.

\bibitem[Breiman(2001)]{breiman_2001}
Breiman, L.
\newblock {Random Forests}.
\newblock {\em Machine Learning} {\bf 2001}, {\em 45},~5--32.

\bibitem[Breiman(1996)]{breiman_1996}
Breiman, L.
\newblock Bagging predictors.
\newblock {\em Machine Learning} {\bf 1996}, {\em 24},~123--140.

\bibitem[Boureau \em{et~al.}(2010)Boureau, Ponce, and LeCun]{boureau_2010}
Boureau, Y.L.; Ponce, J.; LeCun, Y.
\newblock A theoretical analysis of feature pooling in visual recognition.
\newblock  Proceedings of the 27th International Conference on Machine Learning
  (ICML-10),  2010, pp. 111--118.

\bibitem[Ren \em{et~al.}(2015)Ren, He, Girshick, and Sun]{ren_2015}
Ren, S.; He, K.; Girshick, R.; Sun, J.
\newblock Faster r-{CNN}: Towards real-time object detection with region
  proposal networks.
\newblock  Advances in Neural Information Processing Systems,  2015, pp.
  91--99.

\bibitem[He \em{et~al.}(2016)He, Zhang, Ren, and Sun]{he_2016}
He, K.; Zhang, X.; Ren, S.; Sun, J.
\newblock Deep residual learning for image recognition.
\newblock  Proceedings of the IEEE conference on Computer Vision and Pattern
  Recognition,  2016, pp. 770--778.

\bibitem[Bergstra and Bengio(2012)]{bergstra_2012}
Bergstra, J.; Bengio, Y.
\newblock Random search for hyper-parameter optimization.
\newblock {\em Journal of Machine Learning Research} {\bf 2012}, {\em
  13},~281--305.

\bibitem[Hutter \em{et~al.}(2011)Hutter, Hoos, and Leyton-Brown]{hutter_2011}
Hutter, F.; Hoos, H.H.; Leyton-Brown, K.
\newblock Sequential model-based optimization for general algorithm
  configuration.
\newblock  International Conference on Learning and Intelligent Optimization.
  Springer,  2011, pp. 507--523.

\bibitem[Snoek \em{et~al.}(2012)Snoek, Larochelle, and Adams]{snoek_2012}
Snoek, J.; Larochelle, H.; Adams, R.P.
\newblock Practical bayesian optimization of machine learning algorithms.
\newblock  Advances in Neural Information Processing Systems,  2012, pp.
  2951--2959.

\end{thebibliography}

\end{document}